\documentclass{article}

\usepackage[preprint]{neurips_2025}

\usepackage[utf8]{inputenc} 
\usepackage[T1]{fontenc}    
\usepackage{hyperref}       
\usepackage{url}            
\usepackage{booktabs}       
\usepackage{amsfonts}       
\usepackage{nicefrac}       
\usepackage{microtype}      
\usepackage{xcolor}         

 \usepackage{multirow}
\usepackage{graphicx}
\usepackage{algorithm}
\usepackage{algpseudocode}
\usepackage{amsmath}
\usepackage{cleveref}

\title{FreeControl: Efficient, Training-Free Structural Control via One-Step Attention Extraction}

\author{
  \vspace{-25pt}\\
  \textbf{Jiang Lin$^{1,\dag}$,\quad Xinyu Chen$^{1}$,\quad Song Wu$^{2}$, \quad Zhiqiu Zhang$^1$,\quad Jizhi Zhang$^{1}$,\quad Ye Wang$^{4}$,\vspace{3pt}} \\
  \textbf{Qiang Tang$^{3}$,\quad Qian Wang$^{2}$,\quad Jian Yang$^{1}$,\quad Zili Yi$^{1}$\thanks{Corresponding authors: \,\href{mailto:yi@nju.edu.cn}{\color{black}{yi@nju.edu.cn}};\, $\dag$: project lead }}\vspace{3pt} \\
  $^1$Nanjing University, Suzhou, China \\ $^2$ JIUTIAN Research, Beijing, China \\ $^3$University of British Columbia, Vancouver, Canada \\ $^4$Jilin University, Changchun, China\vspace{3pt} \\
  \texttt{\small lin@smail.nju.edu.cn, yi@nju.edu.cn} \\
  \vspace{-30pt}
}

\begin{document}

\maketitle

\begin{figure}[htbp]
    \centering
    \includegraphics[width=1 \linewidth]{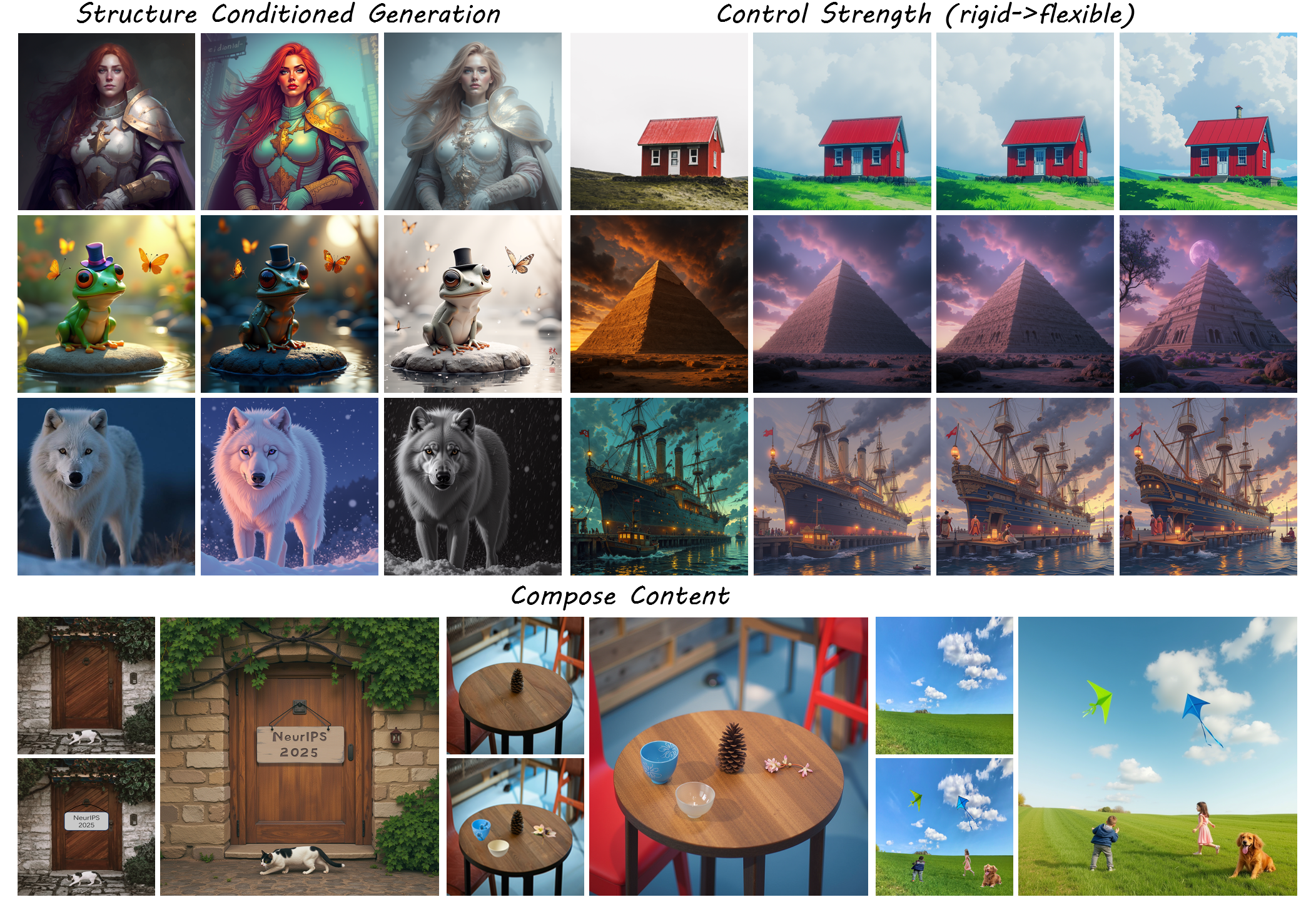}
    \caption{
       FreeControl enables efficient, structure-aware generation from raw image references. Top-left: structure-conditioned generation using reference image on the left. Top-right: Tunable Control strength via adjustable attention injection. Bottom: compositional generation from user-assembled reference images enables intuitive spatial and semantic layout control. 
    }
    \label{fig:teaser}
\end{figure}

\begin{abstract}

Controlling the spatial and semantic structure of diffusion-generated images remains a challenge. Existing methods like ControlNet rely on handcrafted condition maps and retraining, limiting flexibility and generalization. Inversion-based approaches offer stronger alignment but incur high inference cost due to dual-path denoising.
We present \textbf{FreeControl}, a training-free framework for semantic structural control in diffusion models. Unlike prior methods that extract attention across multiple timesteps, FreeControl performs \textit{one-step attention extraction} from a single, optimally chosen key timestep and reuses it throughout denoising.  This enables efficient structural guidance without inversion or retraining.
To further improve quality and stability, we introduce \textit{Latent-Condition Decoupling (LCD)}: a principled separation of the key timestep and the noised latent used in attention extraction. LCD provides finer control over attention quality and eliminates structural artifacts.
FreeControl also supports compositional control via reference images assembled from multiple sources, enabling intuitive scene layout design and stronger prompt alignment. 
FreeControl introduces a new paradigm for test-time control—enabling structurally and semantically aligned, visually coherent generation directly from raw images, with the flexibility for intuitive compositional design and compatibility with modern diffusion models at \textasciitilde 5\% additional cost.

\end{abstract}

\section{Introduction}


While diffusion models ~\cite{ho2020denoising,rombach2022high,saharia2022photorealistic,ramesh2022hierarchical, dhariwal2021diffusion, podell2023sdxl, blattmann2023stable} have revolutionized generative image synthesis, they remain difficult to control. There often lacks intuitive ways to specify what appears where, or how objects should relate spatially and semantically—making them less suitable for tasks like scene layout, object rearrangement, or design prototyping.

A prevalent approach to this challenge involves conditioning generation on external control maps, as exemplified by ControlNet~\cite{zhang2023adding} and T2I-Adapter~\cite{mou2024t2i}. These methods inject spatial guidance via edge maps, depth cues, or segmentation masks. While effective, they depend on handcrafted pre-processing and require separate training for each control type and base model. In particular, ControlNet demands large-scale paired datasets and substantial training resources per condition, making it expensive to scale across modalities or architectures. Moreover, the control signals themselves are limited: Canny edges are often overly rigid and may conflict with prompt semantics, while segmentation-based guidance is restricted by limited category labels, preventing nuanced or open-ended structure control.
In contrast, test-time augmented methods~\cite{song2020denoising,mokady2023null,lin2025inversion,tumanyan2023plug} such as DDIM inversion~\cite{song2020denoising} extract structure from reference images by reconstructing their latent trajectory and injecting features throughout denoising. These techniques incur high inference cost, requiring full or dual-path denoising and considerable memory. 

We propose a training-free, test-time augmented framework for semantic structural control using raw reference images. Our method performs a single additional denoising step at a model-specific key timestep, chosen to extract maximally informative self-attention. This attention matrix captures both spatial structure and semantic intent, and is consistently injected into the main generation process to guide the arrangement and content of the generated image. The strength and scope of guidance are tunable, enabling both flexible layout guidance and strong structural adherence, depending on user intent.
Unlike prior test-time augmentation methods ~\cite{hertz2022prompt,chefer2023attend}, our approach eliminates the need for inversion or reconstruction entirely—removing both the computational burden and architectural complexity of dual-path denoising. With only ~5\% additional cost over baseline inference, it delivers high-quality structural control without retraining, making it directly compatible with fine-tuned~\cite{ruiz2023dreambooth} or LoRA-augmented models~\cite{hu2022lora}.

By collapsing multi-step extraction into a single attention signal, our one-step approach creates a tractable point of analysis—enabling us to systematically study and refine the quality of structural guidance through Latent-Condition Decoupling (LCD). LCD separates the roles of the noised latent and the key timestep, revealing how each factor shapes the extracted structure. This lets us improve alignment quality, reduce artifacts, and offer tunable control over structural granularity—from coarse layout to fine semantic detail.

To support intuitive, layout-aware control beyond segmentation maps or prompt tuning, we introduce a composition-based conditioning strategy. As shown in \cref{fig:teaser}, users can directly assemble reference images by cropping and combining objects from different sources, enabling them to express both spatial layout and semantic intent in a natural visual form, and generate images with content that aligns with their expectations. This flexibility transforms structural and semantic conditioning into a designable interface for high-level scene control. 

In experiments, FreeControl outperforms existing structural control methods in both spatial alignment and visual fidelity, while maintaining high efficiency. Qualitative results further demonstrate its advantage in semantic-level control, producing generations that more faithfully adhere to the intended prompts.

Our contributions are as follows:
\begin{itemize}
 \item We present a training-free, test-time augmented method for semantic structural control from raw reference images, eliminating the need for handcrafted inputs, inversion, or retraining.
\item We propose a one-step attention extraction framework that uses a single denoising step at a key timestep to guide generation, with attention maps injected across layers during inference.
\item We introduce Latent-Condition Decoupling (LCD), a principled method that separates the key timestep from the noised latent in attention extraction, enabling stronger control and improved stability.
\item We introduce a composition-based conditioning approach that allows users to define both spatial layout and semantic intent through assembled reference images, enabling intuitive control beyond segmentation maps or prompt tuning.
\end{itemize}

\section{Related Work}

\textbf{Diffusion Models.}
Diffusion models~\cite{ho2020denoising, song2020denoising, dhariwal2021diffusion, podell2023sdxl} have emerged as a leading framework for high-quality image synthesis, with success across tasks such as text-to-image generation~\cite{rombach2022high, saharia2022photorealistic, ramesh2022hierarchical}, image-to-image translation~\cite{isola2017image,meng2021sdedit, kulikov2024flowedit}, and image editing~\cite{hertz2022prompt, brooks2023instructpix2pix, xu2024inversion, kawar2023imagic, sheynin2024emu}. Foundational models like DDPM~\cite{ho2020denoising} and DDIM~\cite{song2020denoising} introduced the core denoising process, while later developments such as LDM~\cite{rombach2022high}, DiT~\cite{peebles2023scalable}, and SD3~\cite{esser2024scaling} have scaled diffusion to high-resolution, semantically rich generation. The field has also moved from U-Net-based backbones~\cite{ronneberger2015u} to more expressive transformer-based architectures~\cite{vaswani2017attention}.

\textbf{Structural Guidance via Training-Based Conditioning.}
Training-based methods such as ControlNet~\cite{zhang2023adding} and T2I-Adapter~\cite{mou2024t2i} guide spatial structure using condition maps like edges or segmentation. While they achieve strong low-level alignment, they require retraining for each control type and base model—introducing high computational cost and model proliferation. Their reliance on handcrafted inputs also leads to brittle performance when structure maps are noisy or conflict with text prompts. T2I-Adapter is lighter but similarly struggles with complex scenes and still demands per-condition training.
High-level alternatives like GLIGEN~\cite{li2023gligen} and IP-Adapter~\cite{ye2023ip} support layout-aware and visual conditioning using bounding boxes or global image features. However, IP-Adapter still needs base-model-specific training and shows varied quality depending on the dataset. Though these methods improve compositional flexibility, they lack fine-grained structural control—e.g., for object contours or pose—limiting their utility in dense structure transfer tasks.

\textbf{Test-Time Control via Inversion and Attention Reuse.}
Test-time approaches offer another path to structure control. DDIM inversion~\cite{song2020denoising, tumanyan2023plug, liu2024towards} and Null-Text Inversion~\cite{mokady2023null} reconstruct noise latents from reference images, enabling attention reuse for editing. Though effective, they are computationally heavy and depend on prompt alignment. Prompt-to-Prompt~\cite{hertz2022prompt} modifies cross-attention for semantic edits while preserving layout but cannot incorporate visual references.

Plug-and-Play~\cite{tumanyan2023plug} injects image features during inference but offers coarse structure control. It requires dual attention modulation and ResNet backbones, limiting efficiency and compatibility with transformer-based models like DiTs~\cite{peebles2023scalable}. Extracting attention across multiple timesteps also incurs a high computational cost.

Crucially, both inversion-based and inversion-free methods operate under the same assumption: that structure must be extracted progressively across a denoising trajectory. Yet, across timesteps, the role of attention remains consistent—capturing spatial layout and semantic structure. This raises a fundamental question: if attention serves the same purpose at every step, is repeated extraction truly necessary?

\begin{figure}[htbp]
    \centering
    \includegraphics[width=0.85\linewidth]{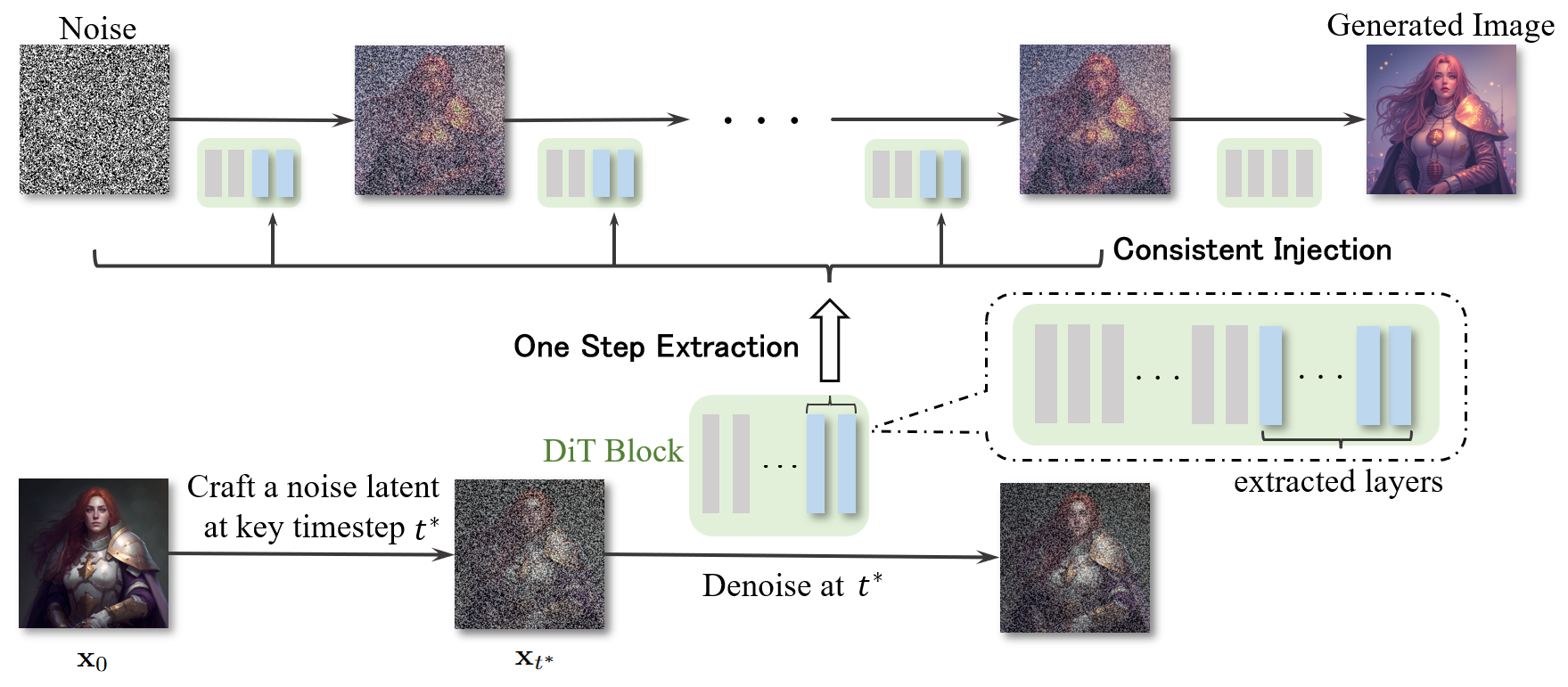}
    \caption{
        The illustration of one-step attention extraction framework. The query attention matrices in the later layers (blue layers) are extracted from a forward-simulated latent at a single key timestep and are injected consistently in the generation process to enable structural guidance.
    }
    \label{fig:comparison_diagram}
\end{figure}

\section{Methods}

\subsection{Semantic Structural Control via One-Step Extraction}
\label{sec:3.1}

\textbf{ One-Step Attention Extraction}
Motivated by the insight that structural information remains conceptually consistent across timesteps, we introduce a \textit{one-step attention extraction strategy} to replace multi-step guidance. As demonstrated in \cref{fig:comparison_diagram}, rather than accumulating structure through repeated attention capture, we extract attention matrices once from a single, designated timestep and reuse them throughout the denoising process. This approach preserves structural alignment while significantly reducing computational overhead.

A key design consideration in this framework is the selection of the optimal timestep \( t^* \) for attention extraction.

We conduct an empirical evaluation across a range of candidate steps and identify the one (661) that yields the strongest structural alignment in the final output. In contrast to inversion-based methods, which require traversing a full reverse denoising trajectory to reach \( t^* \), we adopt a lightweight forward simulation strategy: we apply the forward noise process directly to the reference latent \( \mathbf{x}_0 \) to simulate the noised latent at timestep \( t^* \), bypassing reverse diffusion entirely.
Specifically, we compute the noised latent as:
\begin{equation}
\mathbf{x}_{t^*} = \sigma_{t^*} \cdot \boldsymbol{\epsilon} + (1 - \sigma_{t^*}) \cdot \mathbf{x}_0
\end{equation}
where \( \boldsymbol{\epsilon} \sim \mathcal{N}(0, \mathbf{I}) \) is standard Gaussian noise, and \( \sigma_{t^*} \in [0,1] \) is the timestep-dependent noise scale factor at optimal timestep \( t^* \).
A single denoising step is then applied to \( \mathbf{x}_{t^*} \) to produce intermediate attention maps. From this, we extract the self-attention query matrices \( \mathbf{Q}_{t^*}^{(l)} \) at each transformer layer \( l \).
During generation, these matrices are injected at every timestep \( t \) by replacing the model’s dynamically computed queries:
\begin{equation}
\mathbf{Q}_t^{(l)} \leftarrow \mathbf{Q}_{t^*}^{(l)}
\end{equation}
The key (\( \mathbf{K} \)) and value (\( \mathbf{V} \)) matrices remain dynamically computed from the evolving latent \( \mathbf{x}_t \), preserving responsiveness to the generative context while maintaining consistent structural queries.
This procedure introduces no per-image tuning and requires only a single additional denoising step, making our method highly efficient, training-free, and broadly applicable across diffusion architectures.

\textbf{Layer-Aware Injection for Preserving Appearance Quality.}
While one-step attention extraction provides effective structural control, indiscriminate injection across all layers can degrade visual quality. In particular, injecting structural $\mathbf{Q}$ matrices into early layers of the diffusion transformer interferes with low-level synthesis tasks—such as color, lighting, and texture modeling—often resulting in desaturation, flat textures, or unnatural shading. This occurs because early layers are primarily responsible for fine appearance features, and rigid structural guidance can disrupt their generative flexibility.

In contrast, deeper transformer layers capture higher-level semantic and spatial information, making them more suitable targets for structural injection. To balance structure and appearance, we adopt a layer-aware injection strategy that applies structural $\mathbf{Q}$ matrices only to mid-to-late layers (the blue layers in \cref{fig:comparison_diagram}). This preserves structure alignment while allowing early layers to focus on generating detailed and visually rich content.

\begin{figure}[htbp]
    \centering
    \includegraphics[width=0.9\linewidth]{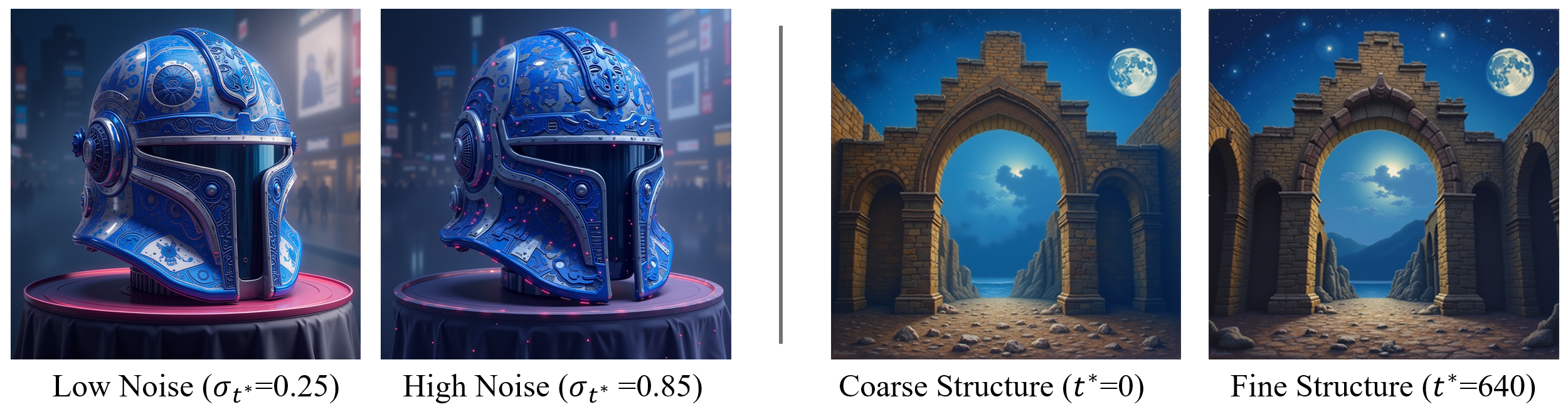}
    \caption{
        Left: Noisy artifacts induced by the noise term. Right: Different granularity of structural control under different key timesteps.
    }
    \label{fig:noise_vs_quality}
\end{figure}

\subsection{Latent-Condition Decoupling (LCD) for Enhanced Attention Quality}
\label{sec:latent_condition_decoupling}

With attention now extracted from a single forward-simulated timestep, we gain a stable and isolated point of intervention for improving control. We introduce \textit{Latent-Condition Decoupling (LCD)} to exploit this opportunity. Rather than varying the key timestep \( t^* \), LCD disentangles the two core influences on attention quality: (1) the noised latent \( \mathbf{x}_{t^*} \) provided as input, and (2) the key timestep passed to the model. By isolating and manipulating these factors independently, we gain deeper insight into how structural guidance arises—and unlock both improved fidelity and fine-grained control.

To isolate the contribution of the noised latent, we fix the key timestep (at previously optimal 661) and vary the construction of \( \mathbf{x}_{t^*} \). As shown in \cref{fig:noise_vs_quality}, we find that latents generated with high noise levels (i.e., large \( \sigma_{t^*} \)) tend to introduce visible noise artifacts such as scattered dots in the final image. These high-noise latents also degrade the attention maps extracted for structural guidance, likely due to the model’s inability to reason clearly over heavily corrupted input. 

Based on this insight, we propose a simplified latent construction that removes the stochastic noise term entirely. 
Rather than performing forward diffusion with sampled noise, we directly construct a scaled $\tilde{\mathbf{x}}$, which serves as a substitute for $\mathbf{x}_{t^*}$.

\begin{equation}
    \tilde{\mathbf{x}} = (1 - \sigma) \cdot \mathbf{x}_0,
\end{equation}

where \( \sigma \) here becomes a tunable scale factor, independent of the key timestep. The removal of the noise term improves the stability of the proposed method. This noise-free latent simulates the amplitude characteristics of an intermediate timestep while preserving the spatial coherence of the original image latent. Empirical evaluation shows that moderate values (e.g., \( \sigma \in [0.25, 0.5] \)) yield the best results.

We then fix the latent input \( \tilde{\mathbf{x}} \) and vary the key timestep passed to the diffusion transformer. As shown in \cref{fig:noise_vs_quality}, the choice of key timestep affects the granularity of structural control. Conditioning with a timestep near zero yields prominent but coarse structure—capturing large shapes and global layout while omitting fine detail. In contrast, using a timestep closer to the original key timestep (e.g., 661) achieves finer structure transfer, preserving contours, texture boundaries, and detailed object shapes.

This observation opens the door to user-driven structural tradeoffs: by adjusting the conditioning timestep, one can control the rigidity of structural guidance. Lower key timesteps provide more compositional flexibility—suitable for creative reinterpretations or stylistic variation—while higher key timesteps enforce stricter alignment with the reference structure. This tunable granularity makes LCD not only a tool for quality improvement, but also a mechanism for interactive control.

\subsection{Unlocking the Full Potential of FreeControl with Compositional Generation}
With the core attention control mechanism in place, FreeControl serves as a flexible framework for structural guidance, enabling users to intuitively define spatial and semantic layout without modifying the model. Rather than relying on hand-crafted segmentation maps or prompt engineering, users can directly control both the content and position of visual elements through image composition.

\textbf{Compositional Reference Images for Semantic Layouts.}
While structural control methods like ControlNet or T2I-Adapter are effective at enforcing edges or spatial layouts, they often struggle to preserve semantic intent—particularly in complex scenes. FreeControl addresses this by enabling compositional reference images, where users can define both *what* should appear and *where* it should appear using direct visual assembly.

For instance, a user can extract an object (e.g., a kite) from a source image using a segmentation tool like SAM~\cite{kirillov2023segment} and paste it onto a new background (e.g., sky). The resulting image encodes both spatial structure and semantic intent, and serves as a direct condition for generation. This “design by composition” approach allows users to guide the layout in a natural, intuitive way—without requiring segmentation maps or prompt engineering. Examples are shown in \cref{fig:teaser}, which comprise cases such as transferring digital text to real writing and scene composition.

To improve robustness when assembling such references,  Gaussian blur could be optionally applied to the compositional image before passing it to the model. This lightweight preprocessing step reduces sharp boundaries and high-frequency noise, helping the model focus on the intended structure while avoiding artifacts from copy-paste seams. 

\section{Experiments}

\subsection{Implementation details}
We conduct all experiments using the FLUX.1-dev~\cite{flux2024} model with the FlowMatchEulerDiscrete scheduler, a timestep range of 1000 to 400, and a guidance scale of 6.5. Quantitative results use 25 denoising steps; 50 steps are used elsewhere for improved visual quality.
The key timestep \( t^* \) is fixed at 661. Attention is extracted once and injected into the last 25 transformer layers of the model’s single transformer block in the quantitative evaluations, and may be reduced elsewhere to demonstrate results of lower structural control. Compositional image generation is disabled unless specifically ablated.
Inference is performed on a single NVIDIA RTX A6000 GPU with 48 GB of memory, and the inference time is measured over 100 runs.

\subsection{Quatitative Comparison} 

{\bf Dataset. }We evaluate on 5,000 images sampled from the COCO 2017~\cite{lin2014microsoft} validation set, resized to 512×512. Each image is paired with its corresponding caption, which is used as the input text prompt for controlled generation.

{\bf Metrics.} We report \textbf{FID} for visual fidelity, \textbf{SSIM} and \textbf{PSNR} for low-level similarity, and \textbf{CLIP-Text Similarity}~\cite{radford2021learning} for semantic alignment between images and prompts. For Canny-conditioned models, we quantify structural fidelity with the \textbf{F1 score} computed between the input Canny edge map and the Canny edge map extracted from each generated image. For depth-conditioned models, we report pixel level accuracy as the \textbf{ mean squared error (MSE)} between the input depth map and the depth map predicted from the generated image.

{\bf Comparison Methods.}We compare FreeControl against five strong baselines: ControlNet~\cite{zhang2023adding}, UniControlNet~\cite{zhao2023uni}, UniControl~\cite{qin2023unicontrol}, ControlNet++~\cite{li2024controlnet++}, and Flux-ControlNet~\cite{XLabs-AI_Flux-ControlNet-Canny, XLabs-AI_Flux-ControlNet-Depth}. The first four baselines are implemented on Stable Diffusion v1.5, while Flux ControlNet is built on FLUX.1-dev\cite{flux2024}. All SD 1.5-based models are run with 20 denoising steps, and Flux-based methods—including FreeControl—use 25 steps, following the respective official configurations.

Note that ControlNet-style methods require pre-processed condition maps (e.g., Canny edge or depth), while FreeControl directly uses the raw image as structural input, with no preprocessing(we also did not count the preprocessing time for the comparison methods in \cref{tab:inferencetime}). The Canny edge is computed with the high threshold set to 200, and the low threshold set to 100.

\textbf{Results.}
Table~\ref{tab:controlled} reports quantitative comparisons across several metrics. FreeControl outperforms all baselines in terms of structural similarity (SSIM and PSNR), while maintaining competitive CLIP-Text alignment with prompt semantics. Compared to ControlNet and UniControl-style methods—which rely on handcrafted edge or depth inputs—our method achieves higher visual fidelity without requiring retraining or specialized condition maps.

In edge-conditioned tasks, FreeControl achieves an F1 score of 0.30 with lower mean squared error (MSE), produces Canny edge results comparable to UniControl and ControlNet++ while using only the raw reference image. Additionally, compared to the iterative extraction, our method performs competitively while exerting significantly less computational resources.

These results demonstrate that FreeControl not only preserves spatial structure and semantic content effectively but also serves as a lightweight, training-free alternative to existing structure-conditioned generation pipelines. For further reference regarding the flexibility and fidelity of our method, we also provide quantitative comparisons that benchmark under different settings in the Appendix.

\begin{table}[t]
\centering
\caption{Comparison with controlled-generation methods. The best scores are in bold, and the second scores are under lined.}
\label{tab:controlled}
\begin{tabular}{lcccccc}
\toprule
\textbf{Configuration} & \textbf{F1 ↑ / MSE ↓} & \textbf{FID ↓} & \textbf{SSIM ↑} & \textbf{PSNR ↑} & \textbf{CLIP-T ↑}\\
\midrule

ControlNet SD1.5 (Canny)~\cite{zhang2023adding}&  0.23 / * & 18.18 & 0.2585 & 10.55  & 0.3083\\
ControlNet++ (Canny) ~\cite{li2024controlnet++} &\underline{0.30} / * & 22.06 & 0.2784 & 10.59 & 0.2986\\
UniControl (Canny)~\cite{qin2023unicontrol} & \textbf{0.35} / * & 21.22 & 0.3714 & 11.66 & 0.3103\\
UniControlNet (Canny)~\cite{zhao2023uni} & 0.26 / * & 17.97 & 0.2783 & 10.59 & \underline{0.3137}\\
FLUX.1-dev ControlNet (Canny)~\cite{XLabs-AI_Flux-ControlNet-Canny}  & 0.16 / * & 27.11 & 0.2515 & 10.65  & 0.3009\\
ControlNet SD1.5 (Depth)~\cite{zhang2023adding}& * / 30.64 & 18.09 & 0.2383 & 10.22  & 0.3107\\
FLUX.1-dev ControlNet (Depth)~\cite{XLabs-AI_Flux-ControlNet-Depth}  & * / 47.04 & 19.27 & 0.1968 & 10.74  & 0.3087\\
ControlNet++ (Depth) ~\cite{li2024controlnet++} & * /  27.79 & 23.23 & 0.2093 & 9.71 & 0.3020\\
UniControl (Depth) ~\cite{qin2023unicontrol} & * / 33.51 & 28.24 & 0.2255 & 10.09 & 0.3105\\
UniControlNet (Depth) ~\cite{zhao2023uni}& * /  34.72 & 22.25 & 0.2038 & 10.12 & \textbf{0.3156}\\
Ours (Iterative Extraction) & \underline{0.30} / \textbf{20.76} & \underline{16.43} & \textbf{0.8078} & \textbf{19.11} & 0.3043\\
Ours (One Step Extraction) & 0.28 / \underline{21.18} & \textbf{15.64} & \underline{0.7564} & \underline{17.49} & 0.3087\\

\bottomrule
\end{tabular}
\end{table}

\begin{figure}[t]
    \centering
    \includegraphics[width=0.95\linewidth]{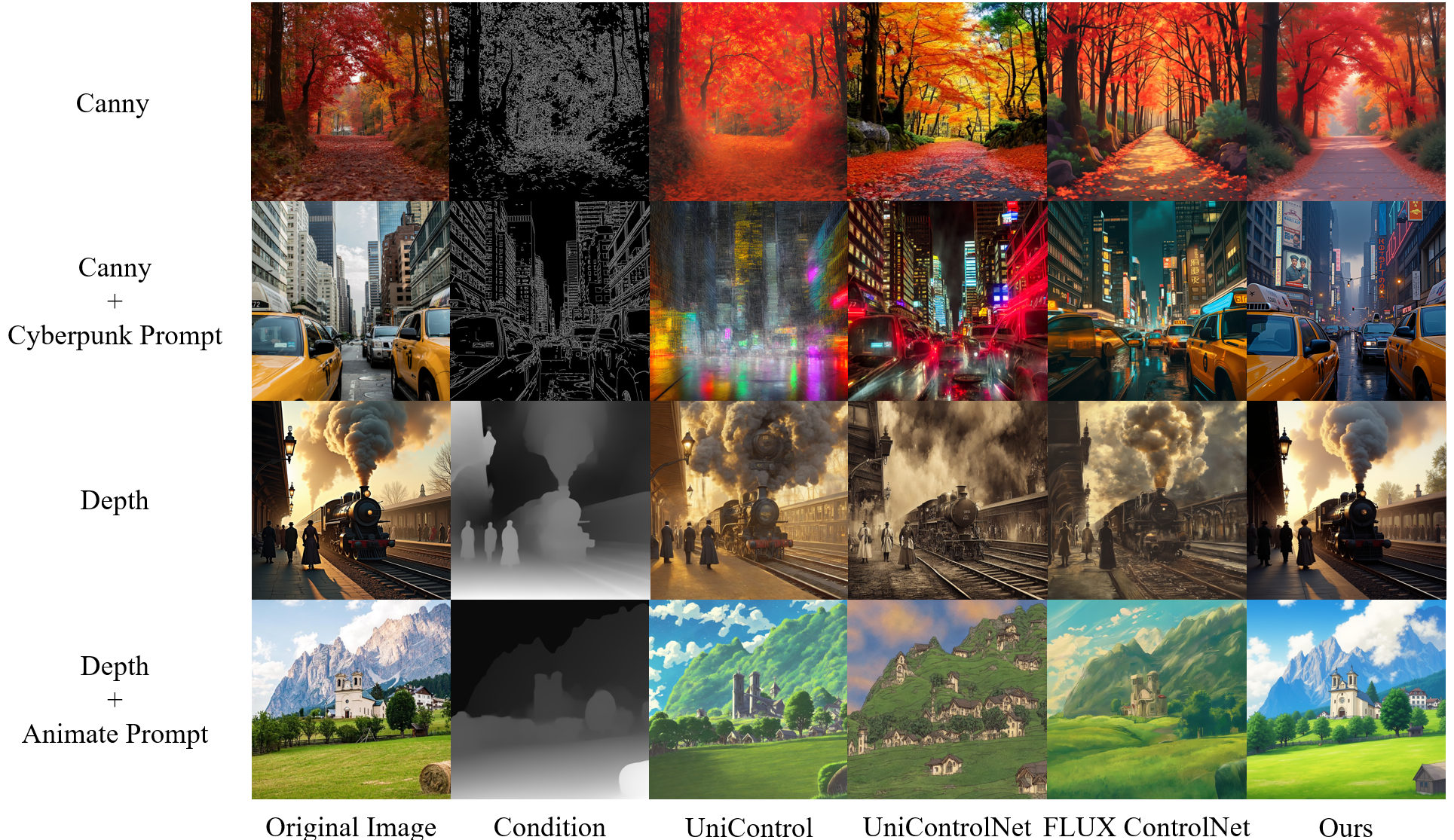}
    \caption{
        Qualitative comparisons on structure-conditioned image generation. Rows 1 and 3 show results where all methods are conditioned using the original caption of the reference image. Rows 2 and 4 present generations under stylized prompts to evaluate each method's ability to generalize beyond the original content.
    }
    \label{fig: qualitative}
    \vspace{-10pt}
\end{figure}

\subsection{Qualitative Results}
We present qualitative comparisons in \cref{fig: qualitative}, where FreeControl is conditioned on raw reference images, while baseline methods rely on preprocessed control signals. Under the original prompt, FreeControl delivers superior structural alignment and visual fidelity. In contrast, comparison methods either fail to align accurately with the intended structure or generate artifacts such as blur or noise, undermining image quality.
We further demonstrate results on representative stylized prompts to evaluate generalization beyond the original setting. FreeControl successfully preserves structural integrity while adapting to new prompts, demonstrating robust guidance under prompt variation. Canny-based methods rigidly adhere to edge maps, often at odds with prompt semantics—resulting in unnatural appearances and ghost artifacts. Depth-based methods suffer from insufficient detail in the control signal, leading to misalignment, semantic drift, and diminished image fidelity.
Overall, the results underscore FreeControl’s ability to maintain consistent structural control and prompt adherence, even when guided by raw image inputs rather than handcrafted control maps.

\subsection{Inference Time}
We benchmark inference speed for FLUX ControlNet (Canny), FLUX ControlNet (Depth), the vanilla FLUX pipeline, and our method. All models are run with 25 denoising steps and produce \(1024 \times 1024\) px outputs on an NVIDIA RTX A6000.
For each pipeline, we fix a single prompt and a single source image—together with its corresponding condition map (canny edges or depth)—and execute the generation 100 times, recording the elapsed time at every run.
Note that we exclude the pre-process time of the comparison methods for a fair comparison.
The aggregate statistics from these 100-run trials are reported in \cref{tab:inferencetime}, and our method, being a test-time augmented method, performs equally efficiently as the training-based methods. Beyond that, the additional memory usage brought by our method is around 1 GB, which is also negligible.

\begin{table}[htbp]
\centering
\caption{Inference time of different methods.}
\label{tab:inferencetime}
\begin{tabular}{lcccc}
\toprule
\textbf{Configuration} & \textbf{Average Inference Time} & Max & Min & Variance\\
\midrule
FLUX Original Pipeline FLUX.1-dev~\cite{flux2024}& 24.89 & 24.98 & 24.18 & 0.0101\\
FLUX.1-dev ControlNet (Canny)~\cite{XLabs-AI_Flux-ControlNet-Canny} & 26.01 & 26.52 & 25.61 & 0.0054\\
FLUX.1-dev ControlNet (Depth)~\cite{XLabs-AI_Flux-ControlNet-Depth} & 26.01 & 26.09 & 25.80 & 0.0034\\
Ours & 26.11 & 26.16 & 25.32 & 0.0117\\
Ours(Iterative Extraction) & 45.16 & 48.09 & 45.01 & 0.1012\\
\bottomrule
\end{tabular}
\end{table}

\subsection{Compatibility with Fine-Tuned or LoRA-Augmented Models}
\begin{figure}[t]
    \centering
    \includegraphics[width=0.9\linewidth]{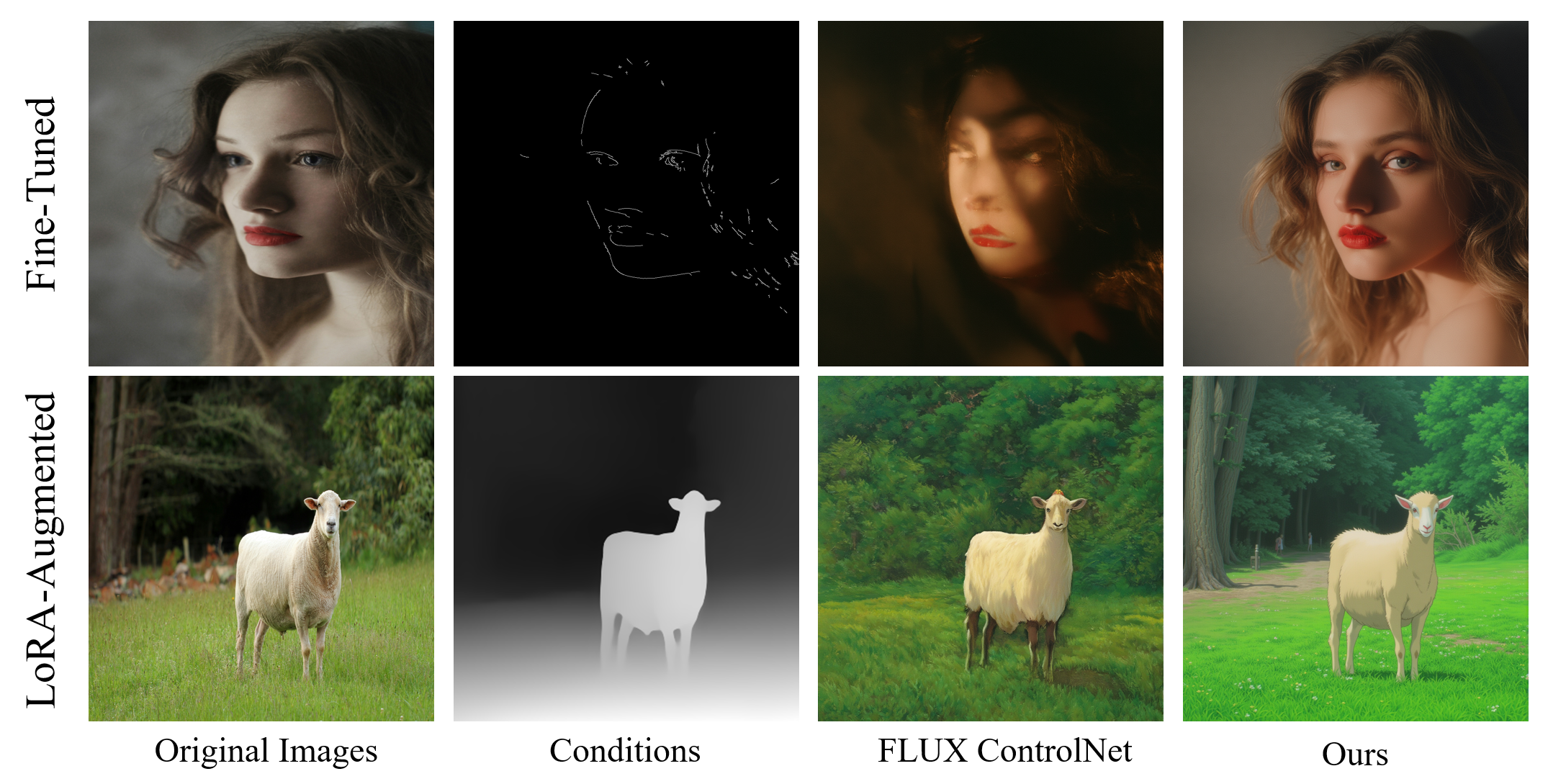}
    \caption{Examples of compatibility with fine-tuned or LoRA-augmented models.}
    \label{fig:ftlora}
\end{figure}

ControlNet~\cite{zhang2023adding} often exhibit limited compatibility with finetuned or augmented via LoRA~\cite{lora}. This instability arises because ControlNet relies heavily on the original backbone's parameters—its control branches are trained jointly with the base model and assume specific internal feature distributions. 
However, our method is not dependent on any specific model architecture or weights, demonstrating strong adaptability across different model variants. 
To validate this, we conduct experiments on both fine-tuned~\cite{awplanet2025awportraitxl} and LoRA-augmented~\cite{openfree2025flux} models, comparing our method with ControlNet FLUX (Canny) and FLUX ControlNet (Depth).
As shown in \cref{fig:ftlora}, our method demonstrates superior compatibility by providing stable results with consistent structure and fine adaptation to the model changes in the community models, whereas ControlNet fails to be compatible with them and produce artifacts and distorted results.  
More results can be found in the Appendix.

\section{Ablation Study}

\subsection{One-Step vs. Iterative Attention Extraction}
To validate the effectiveness of extracting attention from a single timestep, we compare our method against a baseline that mimics iterative attention extraction across multiple denoising steps—similar to inversion-based or reconstruction-based strategies. In this baseline, attention matrices are extracted and injected step-by-step, rather than reused. As shown in \cref{tab:controlled} and \cref{tab:inferencetime}, one-step injection achieves comparable structural fidelity while significantly reducing computational overhead. This result supports our hypothesis that structural information can be captured once and reused without loss of guidance, due to the shared purpose of structural encoding across timesteps.

\begin{figure}[t]
    \centering
    \includegraphics[width=0.85\linewidth]{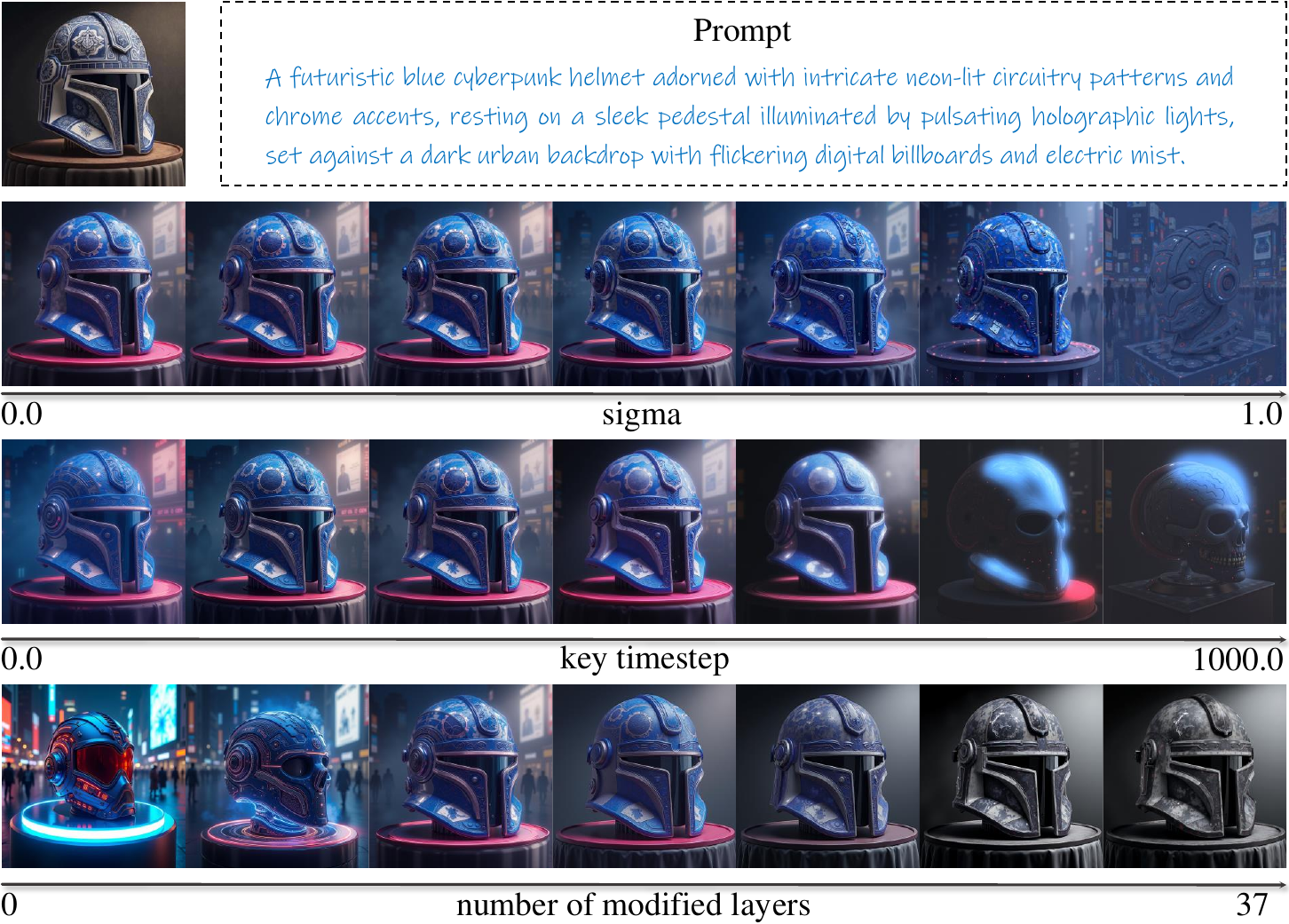}
    \caption{Visual analysis of structural control effects by varying injection depth, sigma, and key timestep in FreeControl.
    }
    \label{fig:layer_choice}
\end{figure}

\subsection{Injection Depth vs. Sigma vs. Key timestep}
The injection depth (number of transformer layers) influences the strength of structural control, while the injection quality (i.e., the content of the attention matrix) determines its focus. To isolate the effects of each factor, we vary it individually while holding others fixed at empirically optimal values. 
As shown in \cref{fig:layer_choice}, varying the injection depth reveals a different trade-off: injecting into fewer transformer layers relaxes structure in less critical regions, improving texture and color fidelity, whereas deeper injection increases rigidity at the cost of visual richness—especially in color saturation.
The choice of key timestep, on the other hand, influences the focus of attention—that is, what kind of structural information is being injected. Earlier timesteps (e.g., \( t = 0 \)) yield more abstract, layout-level attention that allows greater freedom in fine details; later timesteps (e.g., \( t = 661 \)) still capture high-level structure but with greater specificity and finer granularity, resulting in more detailed structural alignment.
The sigma value, in contrast, remains relatively stable at moderate settings, with structural control gradually diminishing as it approaches 1—first affecting fine details, then larger structures.
Based on these observations, we recommend adjusting layer depth and injection range to tune structural strength, while selecting the appropriate key timestep and sigma to steer the level of structural detail captured in the generation.

\subsection{Impact of Layer-Aware Injection}
\begin{figure}
    \centering
    \includegraphics[width=0.9\linewidth]{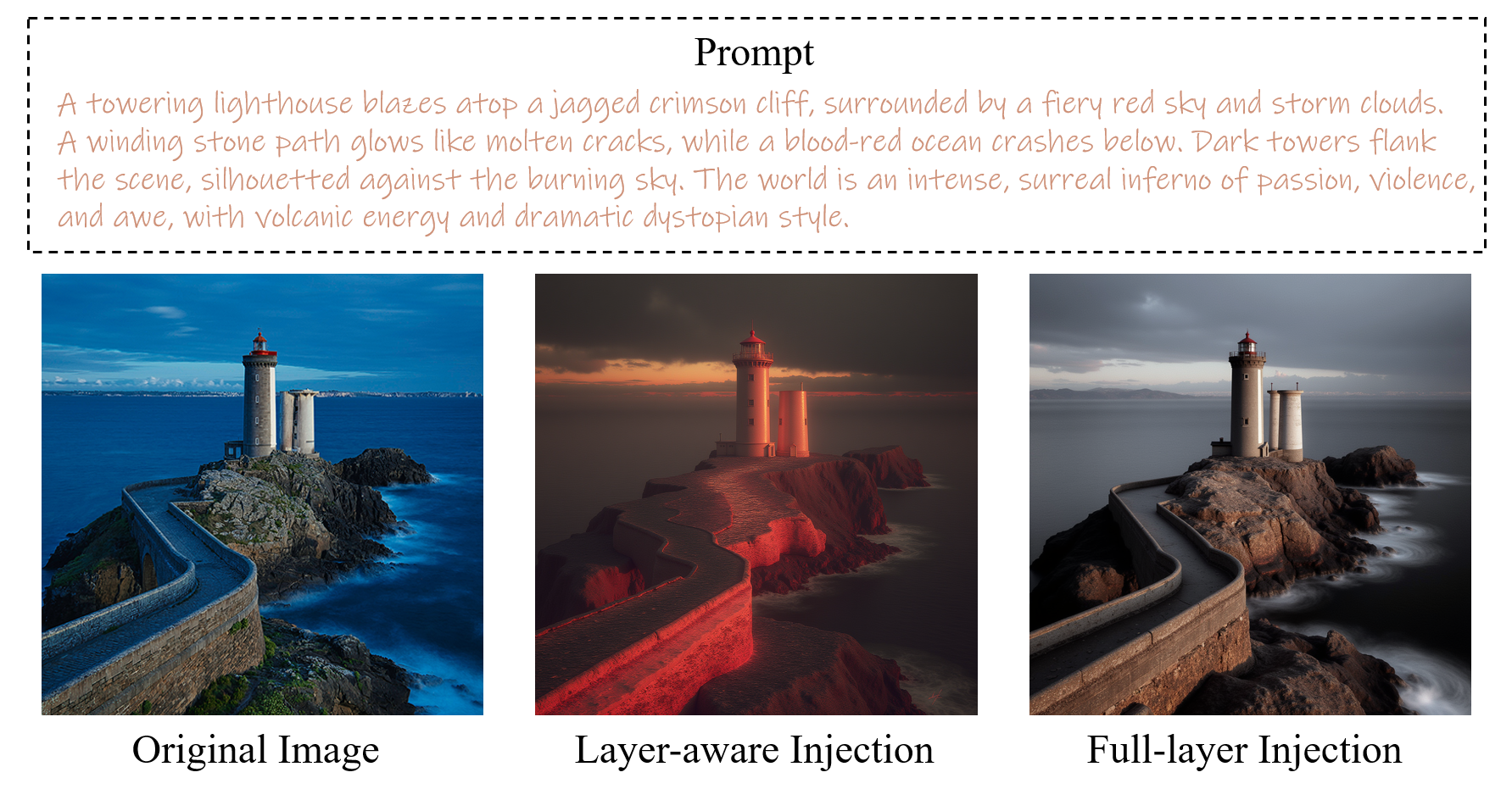}
    \caption{Qualitative Comparison between Layer-Aware Injection and Full-Layer Injection}
    \label{fig:inj_abla}
\end{figure}

As discussioned in \cref{sec:3.1}, 
mid-to-late layer queries encode structural layout rather than raw appearance, aligning better with our structural control objective. 
However, early-layer queries primarily encode low-level appearance statistics, especially color. 
When injected, they conflict with the prompt-conditioned key/value features, which often imply a different color palette. This feature mismatch causes the model to lose chromatic fidelity, leading to muted or grayscale-like outputs.
To verify this, we test prompts with deliberately shifted color themes. 
As shown in \cref{fig:inj_abla}, full-layer injection causes localized color fading in regions where prompt colors diverge from the reference, while our layer-aware injection maintains both structure and visual quality.

\section{Limitations}
While FreeControl is efficient, training-free, and offers strong structural control, it does not support condition maps like edges or segmentation. This limits scenarios where users prefer editing sketches or symbolic inputs. Although compositional image assembly provides flexibility, some use cases may still benefit from explicit support for sparse conditions.

\section{Conclusion}
This paper revisits a central assumption in attention-based structural control for diffusion models: that effective guidance requires multi-step extraction. We show that a single-step extraction—when properly conditioned—can offer strong, reusable structural signals without inversion or retraining.
Our Latent-Condition Decoupling (LCD) reveals that attention quality depends not just on the timestep, but on how the noised latent and conditioning signal are configured. This enables more stable and controllable generation.
Beyond efficiency, FreeControl supports intuitive control by allowing users to compose reference images that express both layout and intent—bridging structure and semantics without relying on edge maps or segmentation masks.
Overall, our findings suggest attention can serve not just as an internal mechanism, but as a practical, tunable approach for control diffusion models.

\section{Acknowledgments}
This work was supported by the National Natural Science Foundation of China (Grant No. 62406134), Jiangsu Provincial Science \& Technology Major Project (Grant No. BG2024042),  the Suzhou Key Technologies Project (Grant No. SYG2024136) and the Nanjing University-China Mobile Communications Group Co. Ltd. Joint Institute.

{
\small
\bibliographystyle{plain}
\bibliography{neurips_2025}
}

\newpage
\section*{NeurIPS Paper Checklist}

\begin{enumerate}

\item {\bf Claims}
    \item[] Question: Do the main claims made in the abstract and introduction accurately reflect the paper's contributions and scope?
    \item[] Answer: \answerYes{} 
    \item[] Justification: I believe they are accurately reflected.

\item {\bf Limitations}
    \item[] Question: Does the paper discuss the limitations of the work performed by the authors?
    \item[] Answer: \answerYes{} 
    \item[] Justification: It is addressed in the Limitations section.

\item {\bf Theory assumptions and proofs}
    \item[] Question: For each theoretical result, does the paper provide the full set of assumptions and a complete (and correct) proof?
    \item[] Answer: \answerNA{} 
    \item[] Justification: This paper does not include theoretical results.

    \item {\bf Experimental result reproducibility}
    \item[] Question: Does the paper fully disclose all the information needed to reproduce the main experimental results of the paper to the extent that it affects the main claims and/or conclusions of the paper (regardless of whether the code and data are provided or not)?
    \item[] Answer: \answerYes{} 
    \item[] Justification: The information provided in the method and experiments section is sufficient to reproduce the results presented in this paper.

\item {\bf Open access to data and code}
    \item[] Question: Does the paper provide open access to the data and code, with sufficient instructions to faithfully reproduce the main experimental results, as described in supplemental material?
    \item[] Answer: \answerNo{} 
    \item[] Justification: The paper has clearly provided the details needed to reproduce the method proposed to the extent to support our claim; however, we are yet unable to provide a properly formulated code regarding the method and all the experiments conducted. This paper will, however, open-source the code regarding its main method upon acceptance.

\item {\bf Experimental setting/details}
    \item[] Question: Does the paper specify all the training and test details (e.g., data splits, hyperparameters, how they were chosen, type of optimizer, etc.) necessary to understand the results?
    \item[] Answer: \answerYes{}  
    \item[] Justification: All the details can be found either in the method section or the implementation details section.

\item {\bf Experiment statistical significance}
    \item[] Question: Does the paper report error bars suitably and correctly defined or other appropriate information about the statistical significance of the experiments?
    \item[] Answer: \answerYes{} 
    \item[] Justification:  Proper information, if needed, is provided to justify the statistical significance of the results, as in \cref{tab:inferencetime}.

\item {\bf Experiments compute resources}
    \item[] Question: For each experiment, does the paper provide sufficient information on the computer resources (type of compute workers, memory, time of execution) needed to reproduce the experiments?
    \item[] Answer: \answerYes{} 
    \item[] Justification: Relevant information is provided in the implementation details, and there is also a table regarding the time of execution\cref{tab:inferencetime}.
    
\item {\bf Code of ethics}
    \item[] Question: Does the research conducted in the paper conform, in every respect, with the NeurIPS Code of Ethics \url{https://neurips.cc/public/EthicsGuidelines}?
    \item[] Answer: \answerYes{} 
    \item[] Justification: The research conforms to the code of Ethics.

\item {\bf Broader impacts}
    \item[] Question: Does the paper discuss both potential positive societal impacts and negative societal impacts of the work performed?
    \item[] Answer: \answerNA{} 
    \item[] Justification: The author believe there is no societal impact of the work performed.
    
\item {\bf Safeguards}
    \item[] Question: Does the paper describe safeguards that have been put in place for responsible release of data or models that have a high risk for misuse (e.g., pretrained language models, image generators, or scraped datasets)?
    \item[] Answer: \answerNA{} 
    \item[] Justification: The paper itself does not release any pretrained models, or datasets.

\item {\bf Licenses for existing assets}
    \item[] Question: Are the creators or original owners of assets (e.g., code, data, models), used in the paper, properly credited and are the license and terms of use explicitly mentioned and properly respected?
    \item[] Answer: \answerYes{} 
    \item[] Justification: The datasets used are properly cited, and the assets are properly referenced if necessary.

\item {\bf New assets}
    \item[] Question: Are new assets introduced in the paper well documented and is the documentation provided alongside the assets?
    \item[] Answer: \answerNA{} 
    \item[] Justification: The paper does not release new assets.

\item {\bf Crowdsourcing and research with human subjects}
    \item[] Question: For crowdsourcing experiments and research with human subjects, does the paper include the full text of instructions given to participants and screenshots, if applicable, as well as details about compensation (if any)? 
    \item[] Answer: \answerNA{} 
    \item[] Justification: The paper does not involve crowdsourcing nor research with human subjects.

\item {\bf Institutional review board (IRB) approvals or equivalent for research with human subjects}
    \item[] Question: Does the paper describe potential risks incurred by study participants, whether such risks were disclosed to the subjects, and whether Institutional Review Board (IRB) approvals (or an equivalent approval/review based on the requirements of your country or institution) were obtained?
    \item[] Answer: \answerNA{} 
    \item[] Justification:  The paper does not involve crowdsourcing nor research with human subjects.

\item {\bf Declaration of LLM usage}
    \item[] Question: Does the paper describe the usage of LLMs if it is an important, original, or non-standard component of the core methods in this research? Note that if the LLM is used only for writing, editing, or formatting purposes and does not impact the core methodology, scientific rigorousness, or originality of the research, declaration is not required.
    \item[] Answer: \answerNA{} 
    \item[] Justification: The core method development in this research does not involve LLMs as any important, original, or non-standard components.

\end{enumerate}
\newpage

\appendix

\section{Quantitative Results on Stylized Prompts (COCO Dataset)}
FreeControl conditions generation directly on the raw image, rather than on derived conditions like edges or depth. This design provides a rich and detailed structural prior, which has proven effective in preserving the original layout and composition during generation. While prior experiments focus on tasks where the target image shares semantic similarity with the reference, real-world use cases often involve prompts that diverge stylistically or conceptually from the original image. To broaden the evaluation scope and assess FreeControl’s robustness in more diverse generative settings, we conduct a benchmark using stylized prompts.

We construct this benchmark on the COCO~\cite{lin2014microsoft} validation set. For each image, we retain the original as the structural reference and generate stylized prompts by combining the original caption with one of five target styles (e.g., Cyberpunk, Vaporwave). These stylized prompts are created using GPT-4o~\cite{gpt4o}, conditioned on both the caption and the style keyword.

This setup introduces a challenging mismatch between the style and the visual structure, making it difficult for models to retain both prompt adherence and structural fidelity. We compare FreeControl against Flux-ControlNet~\cite{zhang2023adding}, which serves as a strong baseline for structure-conditioned generation on the same backbone. As shown in \cref{tab:stylized}, FreeControl consistently preserves structure better (as measured by F1 and MSE) while generating content more semantically aligned with the stylized prompts (via CLIP-T). This demonstrates the strength of our method in transferring structure faithfully even when prompt semantics diverge from the original image.

\section{More Results on Compatibility with Fine-Tuned or LoRA-Augmented Models}
To further support the findings discussed in the main paper, we provide additional qualitative results on fine-tuned and LoRA-augmented diffusion models. Specifically, we evaluate FreeControl and FLUX ControlNet variants on community models that are fine-tuned~\cite{awplanet2025awportraitxl, ifmain2025ultrareal} or LoRA-augmented~\cite{openfree2025flux, prithivMLmods2025canopus}.

\begin{figure}[t]
    \centering
    \includegraphics[width=\linewidth]{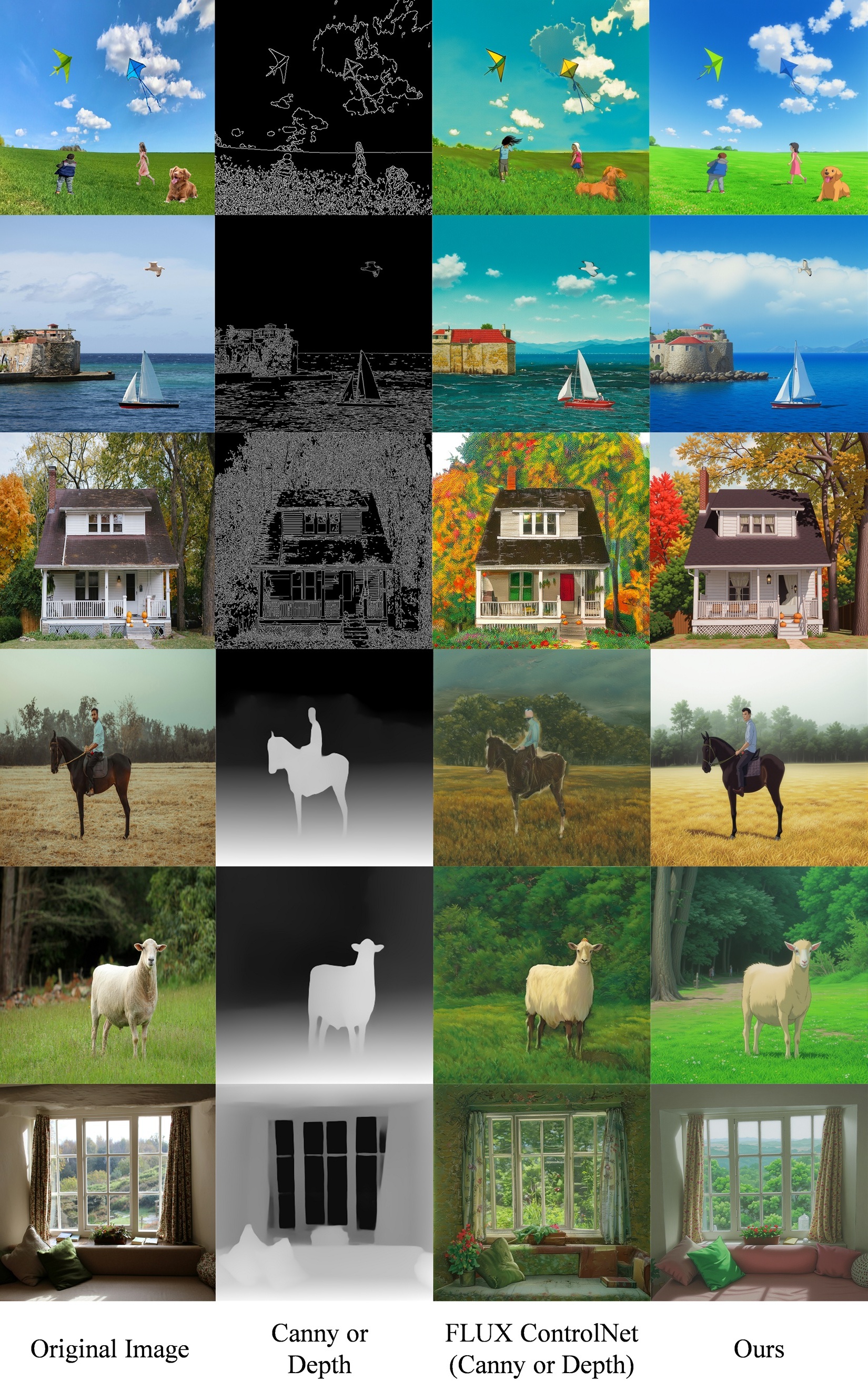}
    \caption{
        Visual results comparing our method and FLUX ControlNet on Lora-Augmented models (Ghibli-Style LoRA).
    }
    \label{fig: lora_comparison1}
\end{figure}

\begin{figure}[t]
    \centering
    \includegraphics[width=\linewidth]{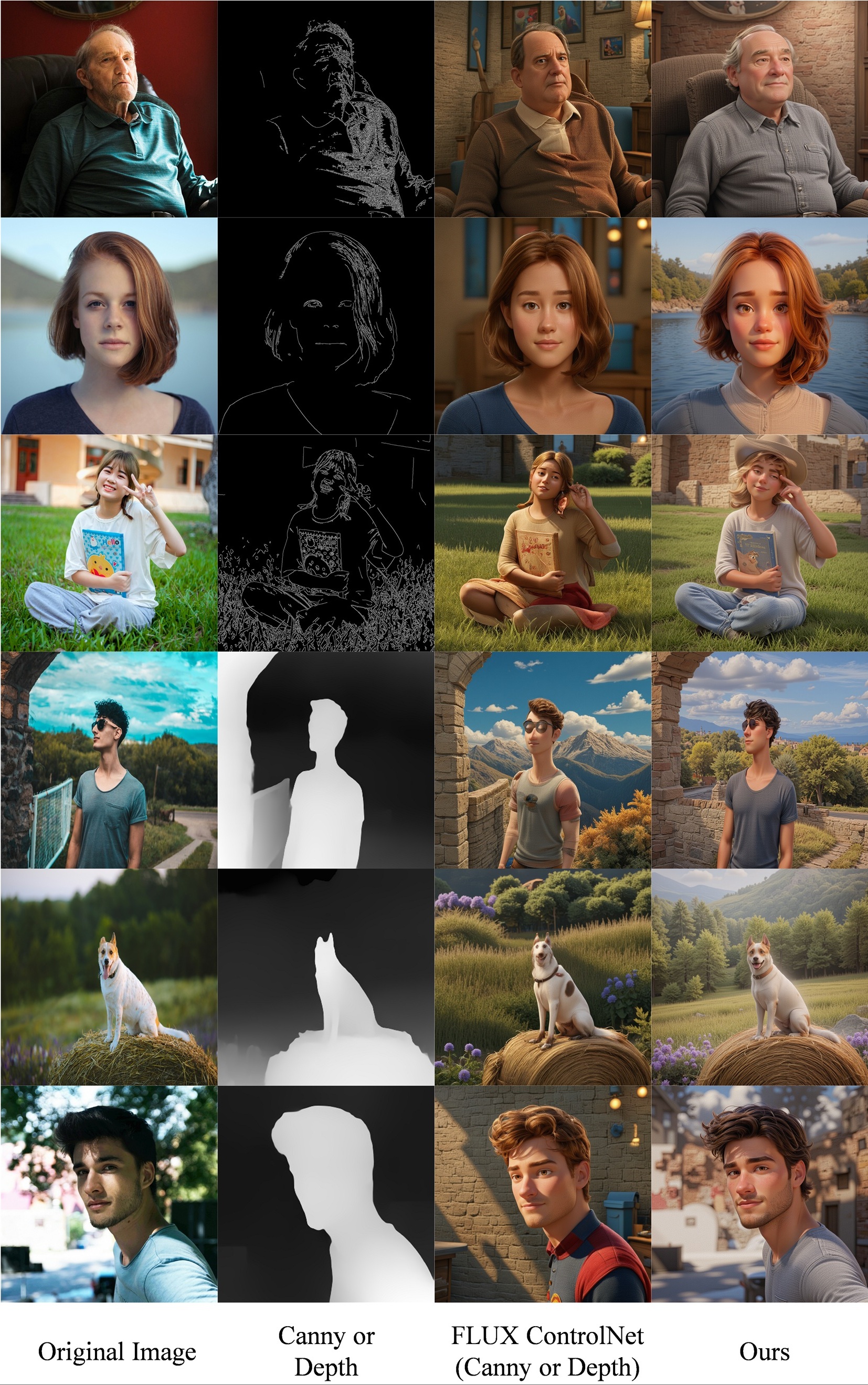}
    \caption{
        Visual results comparing our method and FLUX ControlNet on Lora-Augmented models (Canopus-Pixar-3D-Style LoRA).
    }
    \label{fig: lora_comparison2}
\end{figure}

\begin{figure}[t]
    \centering
    \includegraphics[width=\linewidth]{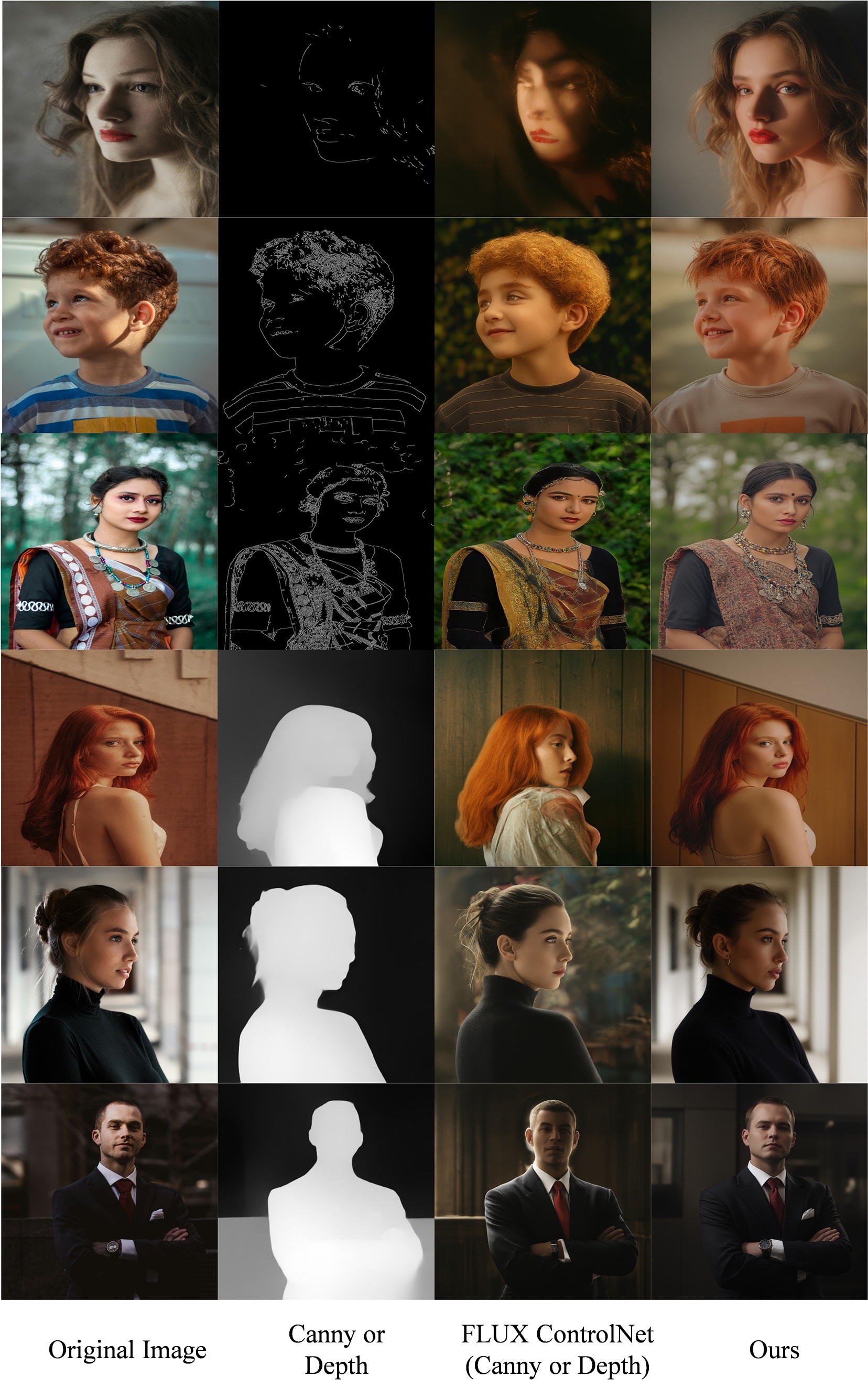}
    \caption{
         Visual results comparing our method and FLUX ControlNet on finetuned models (AWPortrait Fine-Tune).
    }
    \label{fig: finetune_comparison1}
\end{figure}

\begin{figure}[t]
    \centering
    \includegraphics[width=\linewidth]{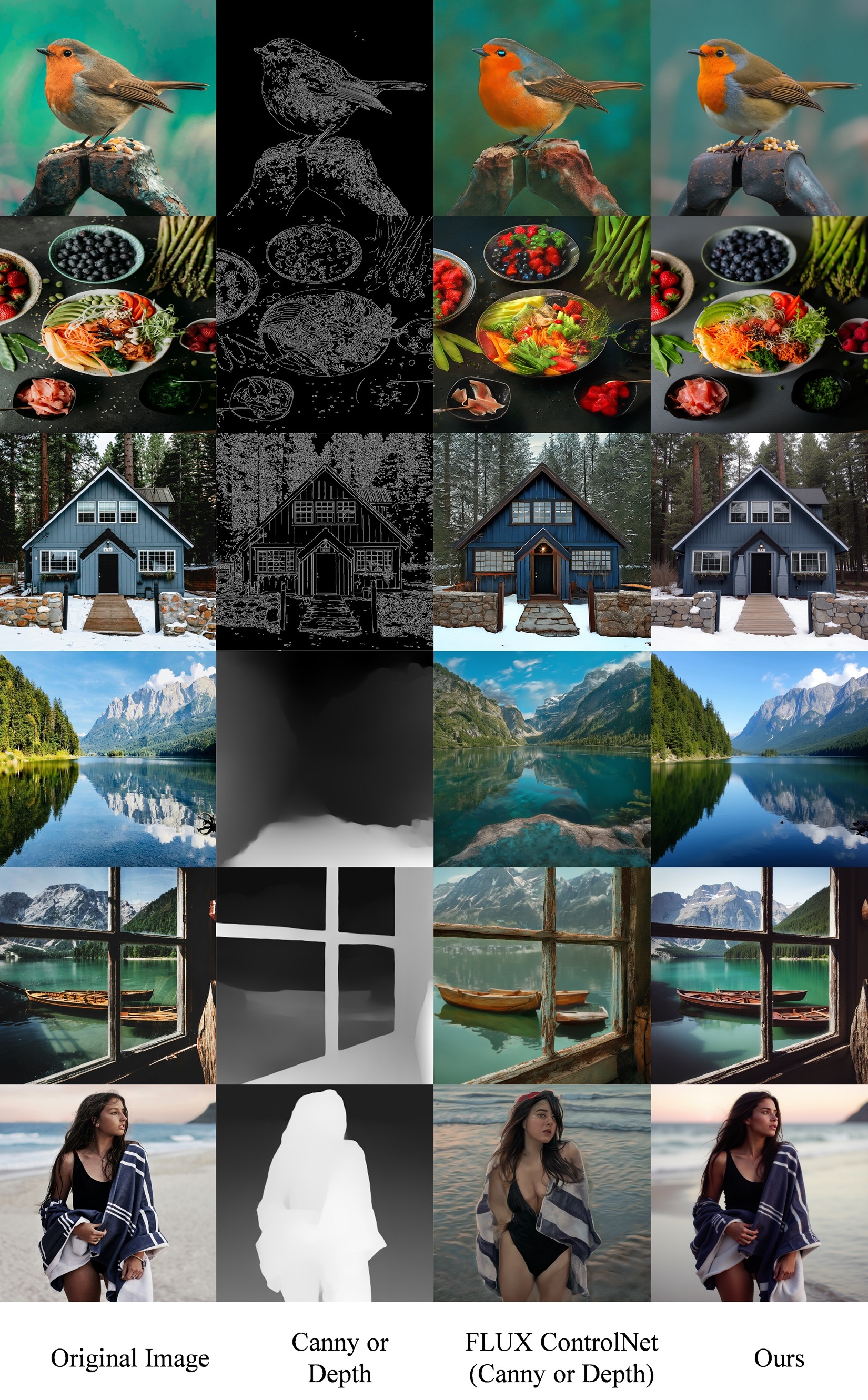}
    \caption{
         Visual results comparing our method and FLUX ControlNet on finetuned models (UltraReal Fine-Tune).
    }
    \label{fig: finetune_comparison2}
\end{figure}

As shown in \cref{fig: lora_comparison1,fig: lora_comparison2,fig: finetune_comparison1,fig: finetune_comparison2}, our method consistently preserves structure and semantic fidelity across diverse model variants, producing stable and visually coherent outputs. In contrast, ControlNet-based approaches exhibit visible artifacts, color shifts, or loss of structural alignment under the same settings.

These results further confirm that FreeControl maintains strong compatibility across both fine-tuned and LoRA-augmented backbones, benefiting from its training-free nature and independence from specific model weights or feature distributions.

\section{Additional Visual Results}
\begin{figure}[htbp]
    \centering
    \includegraphics[width=\linewidth]{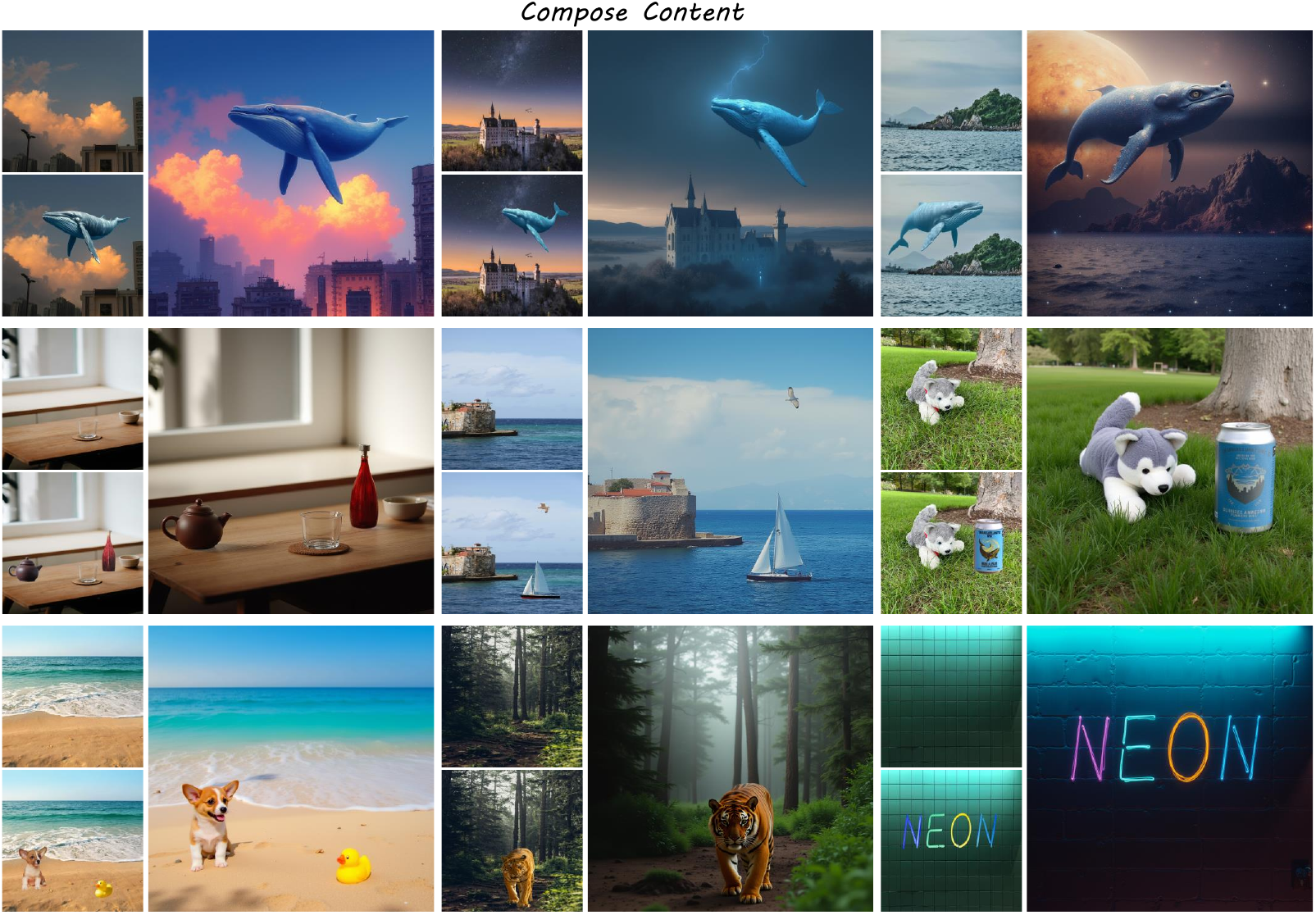}
    \caption{
        More visual results on compositional generation. 
    }
    \label{fig:compositional}
\end{figure}

\begin{figure}[htbp]
    \centering
    \includegraphics[width=\linewidth]{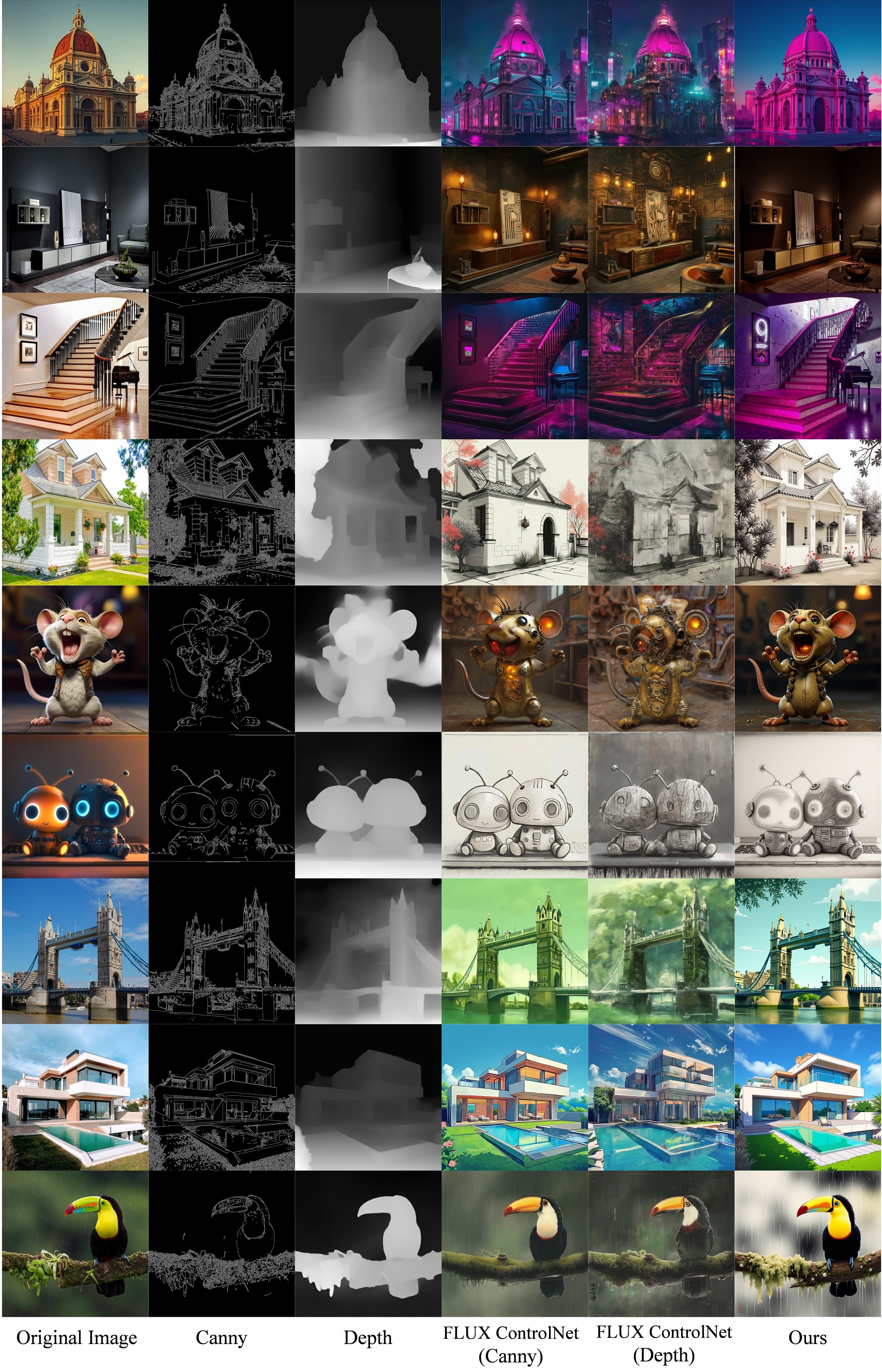}
    \caption{
        Qualitative comparisons on structure-conditioned image generation. 
    }
    \label{fig:baseline_comparison}
\end{figure}

\begin{figure}[htbp]
    \centering
    \includegraphics[width=\linewidth]{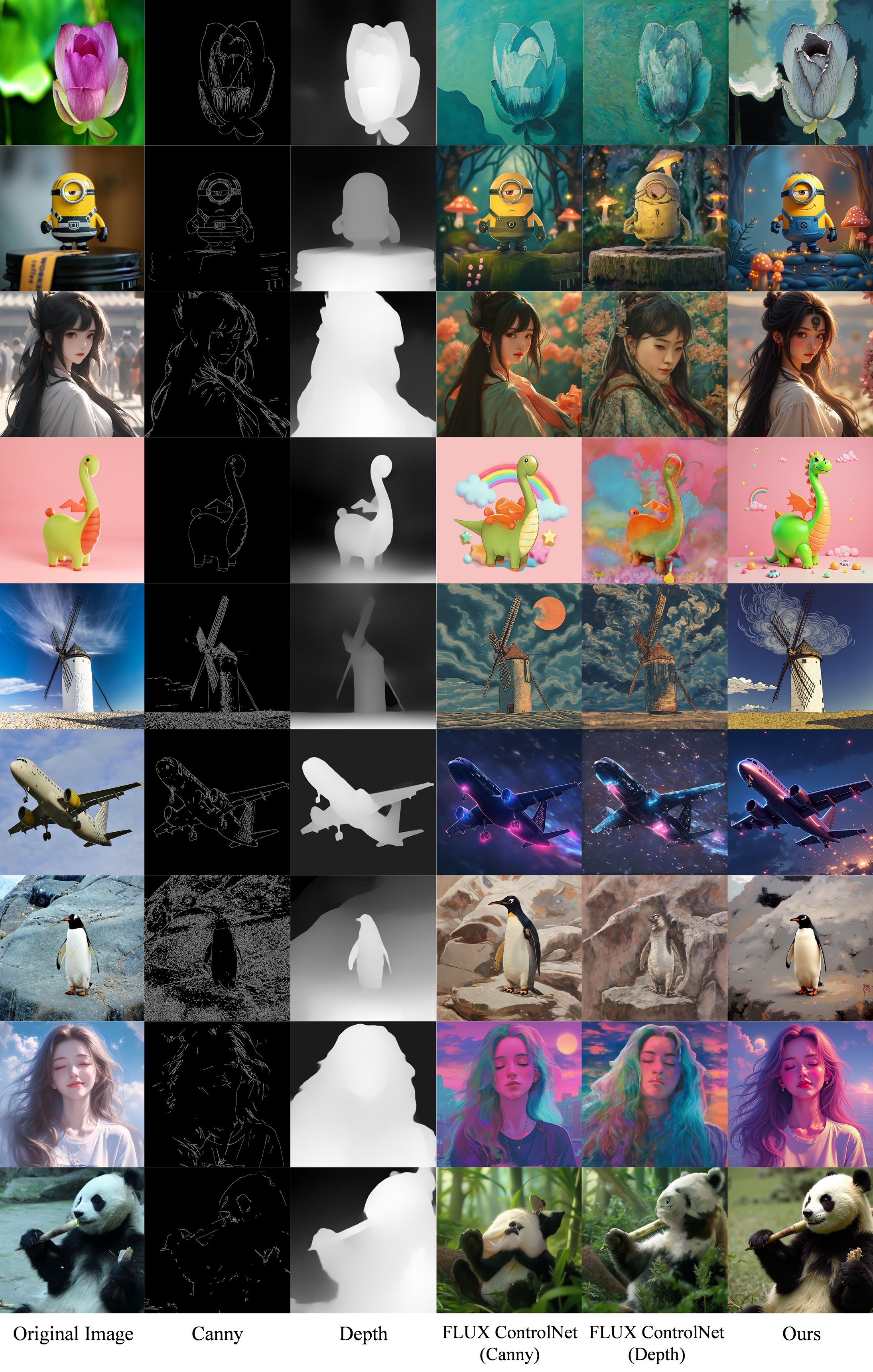}
    \caption{
        Qualitative comparisons on structure-conditioned image generation. 
    }
    \label{fig:baseline_comparison2}
\end{figure}

We provide more visual results for the readers to reference. 
\cref{fig:compositional} showcases additional generation results based on compositional reference images. Users can crop and paste objects into a layout to specify spatial intent, allowing for precise control over the scene’s composition. With a touch of creativity, FreeControl empowers users to bring their imaginative visions to life, generating stunning, dynamic visuals that reflect their unique concepts (like a giant floating whale in the sky).
\cref{fig:baseline_comparison} and \cref{fig:baseline_comparison2} provide further comparisons between FreeControl and FLUX ControlNet.

\begin{table}[htbp]
\centering
\caption{Quantitative evaluation on the COCO validation set using stylized prompts. 
The best scores are in bold.}
\label{tab:stylized}
\begin{tabular}{lcccccc}
\toprule
\textbf{Method}       & \textbf{F1 ↑} & \textbf{MSE ↓} & \textbf{SSIM ↑} & \textbf{PSNR ↑} & \textbf{CLIP-T ↑} \\
\midrule
FLUX ControlNet (Canny)    & 0.19 & N/A & 0.2629 & 9.72  & 0.2646 \\
FLUX ControlNet (Depth)  & N/A &41.41 & 0.2029& 9.98 &   0.2448\\
Ours          & \textbf{0.25} & \textbf{26.24} & \textbf{0.4825}& \textbf{15.07} & \textbf{0.2981}\\
\bottomrule
\end{tabular}
\end{table}

\section{Expanded Baseline Comparison Results}

We have added quantitative comparison experiments with two image editing models, In-Context Edit~\cite{ICEdit} and Taming Rectified Flow~\cite{wang2025tamingrectifiedflowinversion}, to improve the fairness and completeness of the evaluation. 
The test images are consistent with the previous comparison experiments. We use image captions and stylized prompts as text inputs respectively, and the corresponding results are shown in \cref{tab:add_baseline} and \cref{tab:add_baseline_style}.

According to the results, Our method performs comparably to, and often surpasses, Taming Rectified Flow across several metrics. 
ICEdit, built on the FLUX-Fill model~\cite{flux2024} for image inpainting, achieves relatively high similarity metrics (e.g., PSNR) primarily because it keeps all content outside the edited region untouched. However, this strategy limits its ability to satisfy the desired balance between structural control and free content generation. 
As a result, its CLIP-T score is lower, and it often struggles with stability and controllability when following editing instructions.

\begin{table}[htbp]
\centering
\caption{Quantitative Results with image-editing methods on the COCO validation subset. }
\label{tab:add_baseline}
\begin{tabular}{lccccccc}
\toprule
\textbf{Method}              & \textbf{F1 ↑} & \textbf{MSE ↓} & \textbf{SSIM ↑} & \textbf{PSNR ↑} & \textbf{CLIP-T ↑} & \textbf{FID ↓} \\
\midrule
In‑Context Edit             & 0.47 & 17.30 & 0.7781 & 21.34 & 0.3024 & 8.64\\
Taming Rectified Flow           & 0.19 & 28.31& 0.4390 & 16.90 & 0.3164 & 16.35\\
Ours                 & 0.28&21.18 & 0.7564 & 17.49 & 0.3087 & 15.64\\
\bottomrule
\end{tabular}
\end{table}

\vspace{1em}  

\begin{table}[htbp]
\centering
\caption{Quantitative Results with image-editing methods on the COCO validation subset using stylized prompts.}
\label{tab:add_baseline_style}
\begin{tabular}{lcccccc}
\toprule
\textbf{Method}       & \textbf{F1 ↑} & \textbf{MSE ↓} & \textbf{SSIM ↑} & \textbf{PSNR ↑} & \textbf{CLIP-T ↑} \\
\midrule
In‑Context Edit                   & 0.30 & 35.96 & 0.5436 & 14.70  & 0.2543\\
Taming Rectified Flow           & 0.17 &38.49 & 0.4034& 16.37 &   0.2585\\
Ours          & 0.25 & 26.24 & 0.4825& 15.07 & 0.2981\\
\bottomrule
\end{tabular}
\end{table}

\section{Additional Quantitative Ablation Analysis}

We have supplemented more comprehensive ablation studies to justify the choice of key parameters in our method, such as the key timestep $t^*$, numbers of modified transformer layers and $\sigma$. The results are presented in \cref{tab:add_abla}.
Thanks to LCD, by flexibly tuning the hyperparameters, we can achieve structural control of varying strength and granularity, producing stable and controllable results that cater to different user requirements.

\begin{table}[htbp]
\centering
\caption{Ablation results on on the COCO validation subset.
Each entry in the Parameters column indicates the number of modified layers, the key timestep $t^*$, and the $\sigma$, in that order.}
\label{tab:add_abla}
\begin{tabular}{lccccccc}
\toprule
\textbf{Parameters}        & \textbf{F1 ↑} & \textbf{MSE ↓} & \textbf{SSIM ↑} & \textbf{PSNR ↑} & \textbf{CLIP-T ↑} & \textbf{FID ↓}\\
\midrule
20-661-0.25          & 0.27   &   23.01   &  0.5251& 16.61   & 0.3083 & 17.84\\
25-561-0.25         & 0.27   & 22.44    &  0.5232  &16.61   & 0.3077 & 17.42\\
25-661-0.0           & 0.30   & 21.74     &  0.5630  & 17.20  & 0.3061 & 17.78\\
25-661-0.25            & 0.28   &   21.86   &  0.5438  & 16.87   & 0.3048 & 18.00\\
25-661-0.5         & 0.24   &  24.64  & 0.5097 & 16.54   & 0.3072 & 22.51\\
25-761-0.25          & 0.28   &23.47    &  0.5492 & 17.07   & 0.3048 & 21.40\\
30-661-0.25          & 0.30   &   22.20   &  0.5461 & 16.88    &  0.3020 & 20.20\\
\bottomrule
\end{tabular}
\end{table}

\section{Evaluation under Challenging Structural Scenarios}

\subsection{Semantic Entanglement and Object Occlusion}

\begin{figure}
    \centering
    \includegraphics[width=1\linewidth]{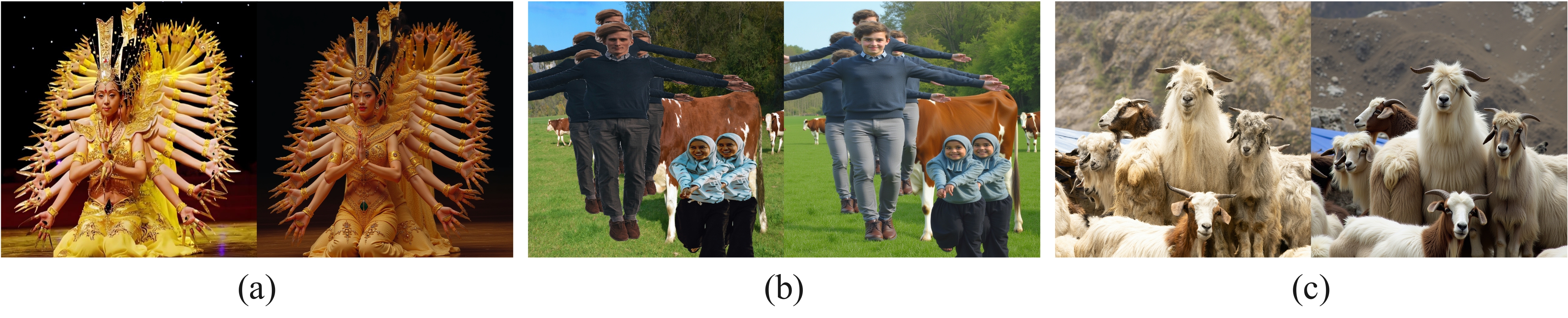}
    \caption{Examples of generated images under Semantic Entanglement and Object Occlusion. For each pair, the image on the left is the original image, and the image on the right is the generated result.}
    \label{fig:Occlusion_and_Entanglement}
\end{figure}

We conduct stress tests on both real and synthetic images featuring semantic entanglement and severe object overlap. Selected results are shown in \cref{fig:Occlusion_and_Entanglement}. For each example, the left image is the original input, and the right image is the generated result. As observed, FreeControl consistently avoids distorted or unrealistic artifacts such as extra limbs or warped body structures. The outputs remain natural-looking and visually coherent, demonstrating strong robustness even under highly challenging compositional scenarios.

\subsection{Preservation of Facial Identity}

\begin{figure}
    \centering
    \includegraphics[width=1\linewidth]{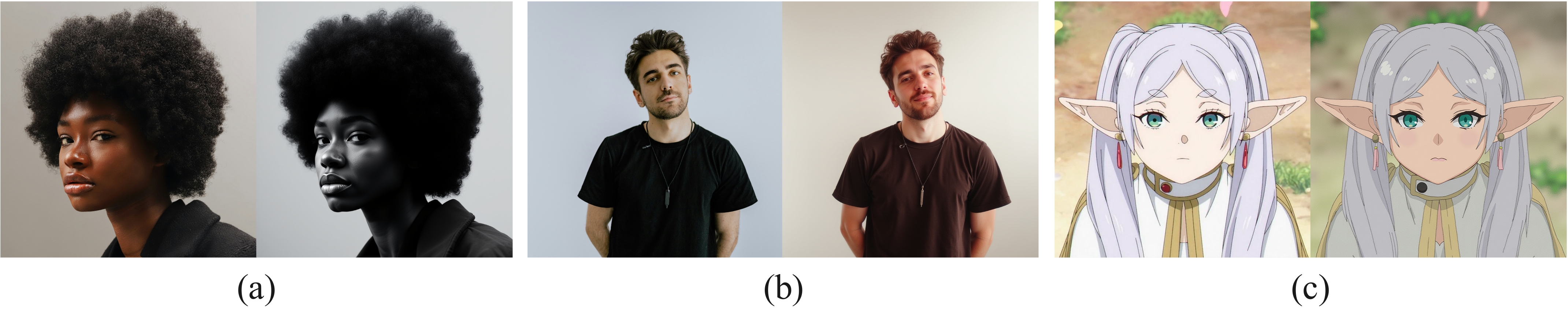}
    \caption{Examples of Facial Identity Control. Adjusting control strength, together with a suitable prompt, enables strong structural preservation of facial features. For each pair, the image on the left is the original image, and the image on the right is the generated result.}
    \label{fig:face id}
\end{figure}

Our method provides flexible control over facial identity preservation, allowing users to adjust the strength of identity retention via hyperparameters. Under non-conflicting text-image guidance, tuning parameters such as the number of modified transformer layers enables strong structural control while preserving facial details, making FreeControl suitable for identity-sensitive tasks.
As shown in \cref{fig:face id}, with higher control strength, our approach demonstrates a strong ability to retain facial identity. 
Conversely, for artistic creation or diversity-oriented generation, relaxing the control allows for slight, intentional changes in facial features, leading to more expressive results.

\section{Applicability to UNet-based Models (i.e., Stable Diffusion)}

\begin{figure}
    \centering
    \includegraphics[width=1\linewidth]{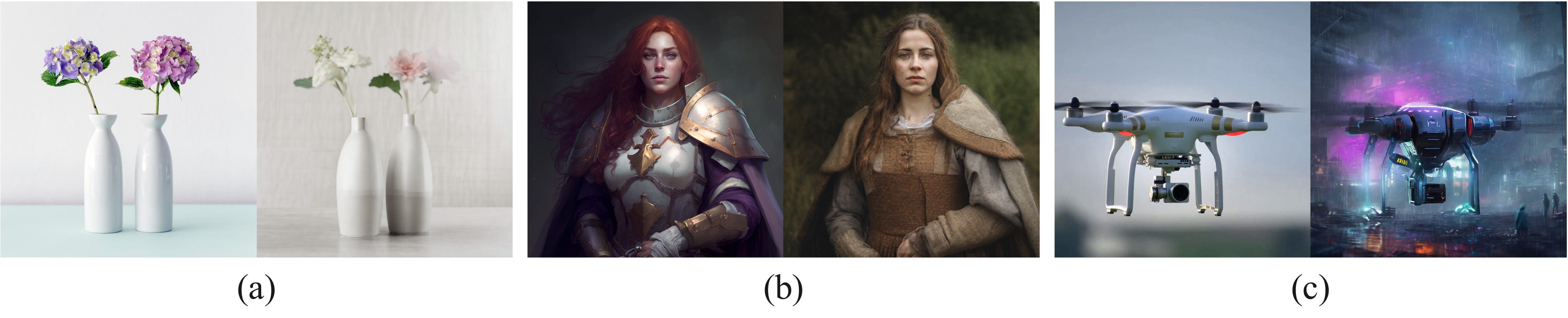}
    \caption{Generation Examples on SDXL model Using Our Method. For each pair, the image on the left is the original image, and the image on the right is the generated result.}
    \label{fig:sdxl_fc}
\end{figure}

\begin{figure}
    \centering
    \includegraphics[width=1\linewidth]{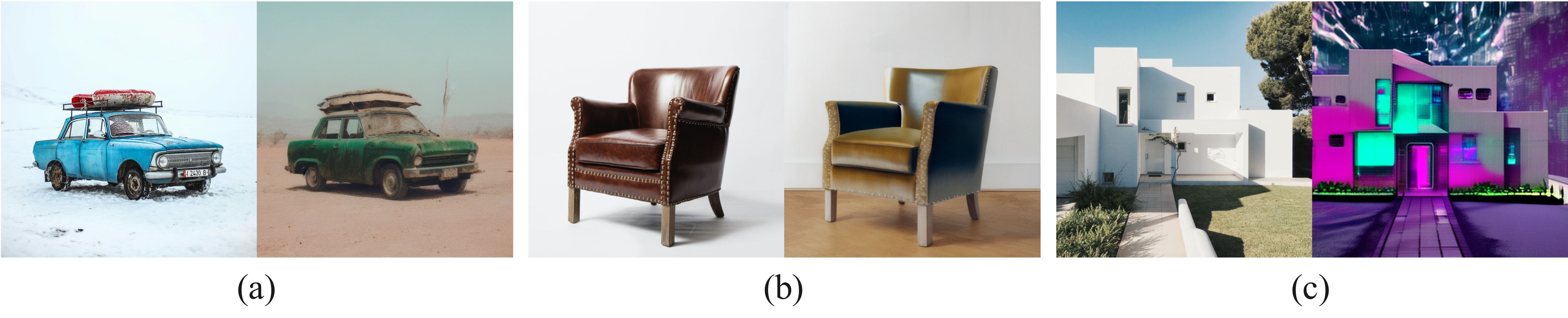}
    \caption{Generation Examples on SD-1.5 model Using Our Method. For each pair, the image on the left is the original image, and the image on the right is the generated result.}
    \label{fig:sd1.5_fc}
\end{figure}

Our method is designed to operate purely at the attention level, making it architecture‑agnostic. 
We have implemented it on UNet–based models (e.g., SD1.5~\cite{rombach2022high}, SDXL~\cite{podell2023sdxl}) and observed strong structural control behavior after only minimal hyperparameters adjustments to fit the model.
Several qualitative examples of structural control are presented in \cref{fig:sdxl_fc} and \cref{fig:sd1.5_fc}. We further note that, due to the inherent capacity limitations of UNet-based models, the degree of controllability can diminish in highly complex scenarios. 
In practice, we find that leveraging FLUX models~\cite{flux2024} yields more stable and visually coherent generations, and we recommend their use when high-fidelity control is desired.

\section{Further Discussion on the Design Space}

\subsection{Independence from ROPE}

The query matrices FreeControl extracts are captured before RoPE~\cite{ROPE} is applied, so the injected queries contain no positional encoding — they are entirely image‑driven. While RoPE still affects key and value during generation, it does not alter what FreeControl injects. Furthermore, FreeControl works identically on U‑Net architectures (which do not use RoPE), showing that structural consistency stems from the extracted queries themselves, not from positional priors.

To directly confirm this point, we ran a controlled test by removing RoPE entirely from the FLUX model. As expected, the base model collapsed into near‑random noise, since it was never trained to operate without positional encoding. Crucially, when we applied FreeControl under the same no‑RoPE setup, the one‑step injection still imposed clear, image‑driven structure on the output. The result looked like “structured noise” faithfully echoing the condition image’s layout — strong evidence that FreeControl’s guidance originates from the injected queries themselves, not from RoPE.

\subsection{Key timestep Choice}

The key timestep fundamentally governs the granularity of structural information that FreeControl can extract. In diffusion models, each denoising step is influenced not only by progressively refined latents but also by a changing timestep input that biases the network toward different levels of detail. Conceptually, the key timestep acts like a focus knob: adjusting it continuously shifts the model’s representational emphasis from global layout patterns to fine-grained textures.

By holding the latent fixed and sweeping only the key timestep, our experiments reveal a natural progression in structural granularity encoded within the query matrices. Both quantitative metrics and visual evidence show a smooth evolution—from coarse shape-level representation toward detailed texture-level encoding. Among all tested values, key timestep (661) emerges as a sweet spot, offering the best trade-off between global consistency and structural precision, making it the most suitable extraction point for query-based control.

\subsection{Layer‑wise Query Matrices Similarity}

We extract query matrices at different depths under two configurations: with LCD and without LCD, where the only difference lies in whether noise is added to $x_0$ during the forward diffusion process (as defined in Sec. 3.1 of main paper). 
We then compare these two sets of query matrices with the query matrices from every timestep of the multi-step extraction variant and compute the cosine similarity, as reported in \cref{tab:q-similarity}.
The layer‑wise similarity shows a clear low‑to‑high trend: shallow layers tend to have lower similarity, while deeper layers converge more. This aligns with our layer‑aware injection choice in Sec. 3.1 — early layers focus more on appearance elements rather than structure, diverging more across timesteps and contributing less to shared structural signals.

\begin{table}[h!]
  \centering
    \caption{Cosine Similarity Between One-Step and Multi-Step Extracted Query Matrices. The ``Global'' row reports the average similarity of query matrices across all layers.}
  \label{tab:q-similarity}
  \begin{tabular}{lcc}
    \toprule
    \multirow{2}{*}{Layer Depth} & \multicolumn{2}{c}{Cosine Similarity} \\
    \cmidrule(r){2-3}
     & w/ LCD & w/o LCD \\
    \midrule
    Early   & 0.5418   & 0.6680 \\
    Mid     & 0.5861  & 0.6853 \\
    Last    & 0.6011  & 0.7638 \\
    Global  & 0.5769 & 0.7063   \\
    \bottomrule
  \end{tabular}
\end{table}

\end{document}